\documentclass{article}
\usepackage{iclr2027_conference,times}
\usepackage[T1]{fontenc}
\usepackage{amsmath,amssymb}
\usepackage{booktabs,multirow,makecell,array,tabularx}
\usepackage{graphicx}
\usepackage{xcolor}
\usepackage{colortbl}
\definecolor{tableheadergray}{HTML}{EBEBEB}
\usepackage{caption}
\usepackage{float}
\usepackage{algorithm}
\usepackage{algpseudocode}
\usepackage{listings}
\usepackage[most]{tcolorbox}
\tcbuselibrary{skins,breakable,listings}
\usepackage{hyperref}
\usepackage{url}

\newcommand{\method}{\textsc{RLHarness}}
\newcommand{\vsbase}[1]{\textsuperscript{+#1}}
\newcommand{\lightrowrule}{\arrayrulecolor{black!18}\hline\arrayrulecolor{black}}
\newcolumntype{L}{>{\raggedright\arraybackslash}X}
\newcolumntype{C}{>{\centering\arraybackslash}X}

\newtcblisting{promptbox}[1][]{
  enhanced,
  breakable,
  listing only,
  colframe=black!72,
  colback=black!3,
  colbacktitle=black!3,
  coltitle=black,
  fonttitle=\small\bfseries\rmfamily,
  title={#1},
  titlerule=0pt,
  arc=1.5mm,
  boxrule=0.7pt,
  top=1.2mm,
  bottom=1.2mm,
  left=2mm,
  right=2mm,
  before skip=0.8em,
  after skip=0.8em,
  listing options={
    basicstyle=\small\ttfamily,
    breaklines=true,
    breakatwhitespace=true,
    columns=fullflexible,
    keepspaces=true,
    showstringspaces=false
  }
}

\title{RLHarness: Co-Evolving Procedural Skills with Reinforcement Learning\\for Long-Horizon Multimodal Reasoning}
\author{%
\normalfont
\parbox{0.96\textwidth}{%
\centering
\mbox{Ziqiao Shang\textsuperscript{\rm 1,2}\textsuperscript{$\dagger$}},
\mbox{Zian Xu\textsuperscript{\rm 1,2}\textsuperscript{$\dagger$}},
\mbox{Ji-Chen Yan\textsuperscript{\rm 3}},
\mbox{Weiming Wu\textsuperscript{\rm 1,2}},
\mbox{Ziyi Jia\textsuperscript{\rm 1,2}},
\mbox{Jie Meng\textsuperscript{\rm 3}},
\mbox{Tao Huang\textsuperscript{\rm 3}},
\mbox{Shan Huang\textsuperscript{\rm 3}},
\mbox{Lan-Zhe Guo\textsuperscript{\rm 1,2}\textsuperscript{*}}\\[0.35em]
\textsuperscript{\rm 1}National Key Laboratory for Novel Software Technology, Nanjing University, Nanjing, China\\
\textsuperscript{\rm 2}School of Intelligence Science and Technology, Nanjing University, Suzhou, China\\
\textsuperscript{\rm 3}Didichuxing Co. Ltd\\
{\small
\textsuperscript{$\dagger$}Equal contribution. \textsuperscript{*}Corresponding author.\\
\textbf{Emails:} shangzq@lamda.nju.edu.cn\hspace{1em}241880148@smail.nju.edu.cn\hspace{1em}yanjichen@didiglobal.com\\
wuwm23@smail.nju.edu.cn\hspace{1em}jiazy@smail.nju.edu.cn\hspace{1em}jmengjie@didiglobal.com\\
alexhuangtao@didiglobal.com\hspace{1em}lattehuang@didiglobal.com\hspace{1em}guolz@nju.edu.cn\\
\textbf{Code:} \url{https://github.com/Ziqiao-Shang/RLHarness}
}%
}%
}
\iclrfinalcopy

\begin{document}
\raggedbottom
\maketitle

\begin{abstract}
Multimodal reasoning requires models to preserve visual evidence through long decision chains while selecting appropriate procedures across diverse scenarios and rules. When learning is guided only by terminal verifiers, reinforcement learning (RL) reveals whether a final answer is correct but not how it should be produced. The policy must therefore discover reusable reasoning procedures while learning to execute them, creating a program cold-start problem. Skills can externalize successful procedures, reduce repeated exploration, and provide inspectable guidance. However, a fixed Skill Bank assumes that this guidance remains compatible with an evolving policy, while updating Skills alone can leave their triggers, execution protocols, and demonstrations stale or mutually inconsistent. We introduce \method{}, which organizes Skills, selection and execution protocols, few-shot demonstrations, and task contracts into a unified, versioned Harness and alternates Harness evolution with policy learning. An Exploration--Distillation Harness builds the initial Harness and version-aligned verified traces for SFT and DAPO I. After the first RL block, a Post-RL Reconstruction Harness rebuilds Skills, protocols, and demonstrations from fresh success--failure rollouts, and DAPO II adapts the policy to the reconstructed program. \method{} improves Accuracy from 16.25\%/27.50\% to 62.00\%/50.00\% on MetroMap-lite/TravelMap-lite and raises F1 score from 37.13\%/45.50\% to 65.81\%/65.51\% on Fee-VL/Cancel-VL. All four tasks achieve their best results only after reconstruction and DAPO II, showing that an evolving Harness complements RL by continually updating the external program that the policy learns to execute.
\end{abstract}

\begin{figure}[ht]
\vspace{-3pt}
\centering
\includegraphics[width=0.99\textwidth]{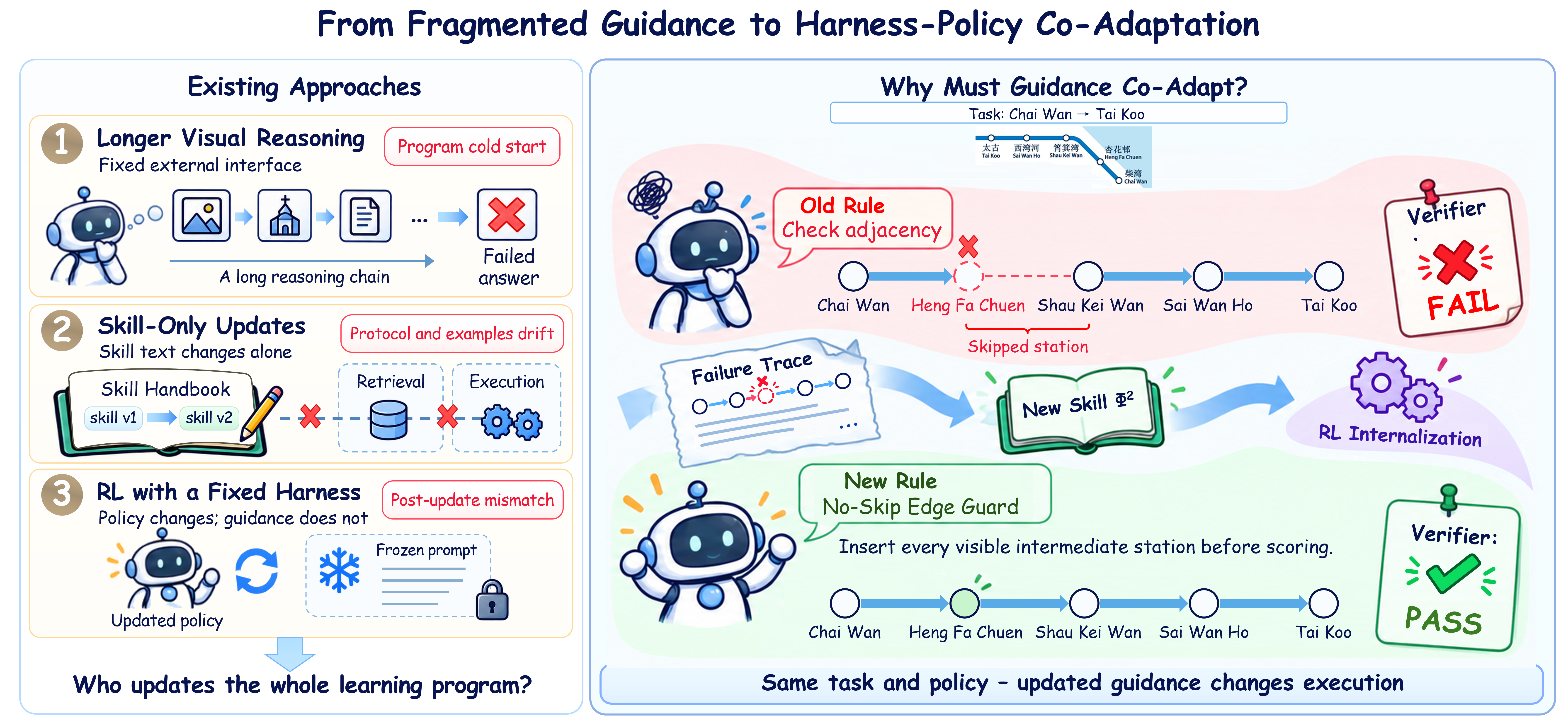}
\setlength{\abovecaptionskip}{3pt}
\setlength{\belowcaptionskip}{0pt}
\caption{Why guidance and policy must co-adapt. \emph{Left:} Longer reasoning with a fixed interface, Skill-only updates, and RL with a frozen Harness leave complementary failure modes: program cold start, drift between Skills and their execution protocol, and mismatch after the policy changes. \emph{Right:} On the same Chai Wan--Tai Koo task, an adjacency-only rule permits a shortcut that skips the visible intermediate station Heng Fa Chuen and fails verification. The resulting failure trace motivates a No-Skip Edge Guard in the reconstructed Harness $\Phi^2$; subsequent RL internalizes the revised execution rule, yielding the complete route and a verifier-accepted answer.}
\label{fig:rlharness-overview}
\vspace{-4pt}
\end{figure}

\section{Introduction}

Complex multimodal reasoning requires more than recognizing isolated visual facts. A model must organize visual and structured evidence into a reliable execution process and preserve that evidence across dependent decisions. In map planning, one omitted station can invalidate every later route and cost calculation~\citep{shang2026maptab}; in rule-intensive judgment, one misplaced condition can change several related labels. These tasks therefore test both long-chain evidence consistency and reliable procedure selection across cases~\citep{wang2026scalelogic}.

Verifiable rewards are useful here because they can score final outputs, and process rewards can sometimes identify partially correct steps~\citep{yuan2026vpr}. But such feedback still need not say how to organize evidence, order the steps, or repair a reusable rule. The policy must explore for an effective program while learning to execute it, which creates a \emph{program cold start}. External Skills reduce this burden by turning prior successes into explicit, reusable guidance instead of forcing each rollout to rediscover the same procedure~\citep{wang2025sage,xia2026skillrl}.

The central difficulty is that the value of this program depends on the policy that executes it. Initial Skills may come from a teacher or early exploration, whereas the eventual user is a student changed by SFT and RL. As the policy changes, some steps become stable, errors move to other parts of the process, the old level of detail or ordering may become unsuitable, and useful rules may need sharper triggers. The initial program is therefore not a timeless specification; it is a hypothesis about what the current learner needs.

For this reason, the update unit cannot be the Skill Bank alone. Suppose a new Skill says to check whether a candidate edge skips an intermediate station, while the protocol still asks the model to compute route cost first and the demonstrations still show one-pass map reading. The Skill, its execution order, and its examples then give inconsistent guidance. We therefore treat the Skill Bank $B$, selection and execution protocol $P$, and few-shot demonstrations $E$ as interdependent parts of a versioned \emph{Harness}; Skills state \emph{what} to do, the protocol states \emph{when and in what order}, and examples show how those rules apply to an input. We jointly revise $B/P/E$ while keeping input assembly, the output contract, and the verifier fixed. Recent studies already co-adapt Skills or broader Harnesses with policies~\citep{shi2026skill1,he2026reskill,chen2026harnessforge,chen2026coharness}; the next section compares their update units and learning settings in detail. Our narrower question is: \emph{for static multimodal reasoning, can current-policy successes and failures be used to reconstruct these interdependent program components, and does that reconstruction improve subsequent RL?}

\method{} answers this question with four stages. \textbf{Harness I} builds the initial program and generates verified supervision that follows the same program. \textbf{Policy Learning I} uses SFT and the first RL block to teach the student to execute it and to collect evidence from the current policy. \textbf{Harness II} contrasts the new successes and failures and reconstructs $B/P/E$ as one version. \textbf{Policy Learning II} then teaches the student to use the reconstructed guidance:
\[
(\mathcal D,\Psi^0)\xrightarrow{\mathcal H_{\mathrm{ED}}}(\Phi^1,\mathcal G^1)
\xrightarrow{\mathrm{SFT+DAPO\ I}}\theta^1
\xrightarrow{\mathcal H_{\mathrm{PR}}(\mathcal R_{\theta^1})}\Phi^2
\xrightarrow{\mathrm{DAPO\ II}}\theta^2.
\]

The experiments separate reconstruction from additional training. Under the same subsequent RL budget, the reconstructed Harness outperforms continuing with the old Harness on all four tasks. Directly replacing the Harness without policy adaptation, however, does not always help. Together, these results support the combination of program reconstruction and subsequent policy adaptation, within the covered MetroMap-lite, TravelMap-lite, Fee-VL, and Cancel-VL settings.

Our contributions are threefold:
\begin{enumerate}
    \item \textbf{Method design.} We present a staged framework that uses current-policy trajectories to jointly reconstruct Skills, their invocation protocol, and their demonstrations.
    \item \textbf{Controlled evaluation.} We separate the immediate effect of replacing a Harness, the effect of subsequent policy adaptation, and the effect of spending the same additional RL budget under the old Harness.
    \item \textbf{Empirical analysis.} On map planning and rule-judgment tasks, we study the gains from program reconstruction, the roles of its components, and limited cross-task transfer.
\end{enumerate}

\section{Related Work}

\subsection{Multimodal reasoning under verifiable feedback}
Multimodal reasoning research improves visual grounding, trajectory supervision, and feedback granularity. Perception-R1 discourages reasoning that bypasses images, Thinking with Images incorporates image transformations into reasoning, and MapTab shows how local cross-modal errors propagate through planning chains~\citep{xiao2025perceptionr1,su2025thinkingimages,shang2026maptab}. TraceR1 trains executable trajectories, Prioritizing the Best distinguishes answer-correct traces by evidence quality, ScaleLogic studies the cost of long-horizon training, and Verifiable Process Rewards supplies step-level feedback~\citep{liang2026tracer1,jia2026groupwise,wang2026scalelogic,yuan2026vpr}. Rather than introducing another reward, we keep the task verifier fixed and use its success--failure traces to revise the external program for subsequent execution.

\subsection{External Skills and Skill--policy learning}
External Skill libraries keep procedural knowledge editable outside model weights, supporting rollout guidance, revision, reuse, retrieval, and maintenance~\citep{wang2025sage,xia2026skillrl,yang2026autoskill,zhou2026mementoskills,pu2026skillops,lin2026muse}. Recent Skill--RL methods connect this knowledge to policy learning but optimize different objects. ReSkill compares Skill versions within GRPO, Co-Evolving Skill Generation and Policy Optimization estimates marginal utility before storage, and SkillForge refines Skills through interaction~\citep{he2026reskill,zhang2026skillpolicy,yang2026skillforge}. These methods mainly update individual Skills or libraries; we instead atomically update the Skill Bank, invocation protocol, and demonstrations as one version. Skill-R1 trains a Skill generator while freezing the task LLM, and Skill1 learns Skill selection, execution, and distillation in one policy~\citep{vishe2026skillr1,shi2026skill1}; we reconstruct the external program from current-policy traces and adapt the student in a separate RL block. SkillForge and SPyCE also involve environment actions or multimodal tools~\citep{yang2026skillforge,zhang2026spyce}, while our static-task setting fixes input assembly, output contracts, and verifiers. RLHarness therefore focuses on version-aligned reconstruction of $B/P/E$ from current-policy traces and subsequent student adaptation under fixed task interfaces.

\subsection{Harness optimization and model adaptation}
Harness research expands the editable object from Skills to the surrounding execution system. The Natural-Language Agent Harnesses framework externalizes control logic, Meta-Harness searches over context-management code, Retrospective Harness Optimization learns from execution history, and Agent Lightning supplies infrastructure that separates agent execution from RL optimization~\citep{pan2026nlah,lee2026metaharness,pan2026rho,he2026agentlightning}. HarnessForge, Co-Harness, and WHALE more directly connect Harness changes with policy or weight updates, while SafeEvolve and On-Policy Correction emphasize safety and compatibility during co-adaptation~\citep{chen2026harnessforge,chen2026coharness,kim2026whale,mao2026safeevolve,yu2026onpolicycorrection}. Our focus is different: under fixed inputs, outputs, and verifiers, RLHarness reconstructs $B/P/E$ from a changed policy and then trains a small model to internalize the revised program.

\section{Method: RLHarness}

\subsection{Problem Formulation}

For each multimodal task domain $d$, an example $(x,y)\in\mathcal D_d$ has input $x=(v,s,q)$, where $v$ denotes visual evidence, $s$ denotes structured information, and $q$ is the task instruction; the target $y$ can be checked by a task verifier. Given a task-specific Harness $\Phi_d$, the parametric policy $f_\theta$ predicts $\hat y=f_\theta(x;\Phi_d)$. Our common objective is
\[
\max_{\theta,\Phi_d}\;J_d=\mathbb E_{(x,y)\sim\mathcal D_d}\left[V_d\!\left(f_\theta(x;\Phi_d),y\right)\right].
\]

\paragraph{Map-route planning.}
The input $x_d^{\mathrm{map}}=(m,t,q)$ contains a map image $m$, structured information $t$ such as the vertex table and edge weights, and route requirements $q$. The output is a complete ordered path $y_d^{\mathrm{map}}=\pi=(u_0,\ldots,u_L)$. The verifier checks the endpoints, station order, constraints, and the full route. An early error in graph recovery or local route selection can therefore propagate through later steps and invalidate the entire path.

\paragraph{VL rule adjudication.}
For VL tasks, $x_d^{\mathrm{VL}}=(i,r,q)$ combines one task image $i$, a structured order record $r$, and an instruction $q$. The output $y_d^{\mathrm{VL}}$ contains a fact-label vector $\mathbf z\in\{0,1\}^{K_d}$, with $K_d=9$ for Fee-VL and $K_d=7$ for Cancel-VL; Cancel-VL additionally predicts responsibility class $c$ for resolved samples. The verifier checks each fact label and, for Cancel-VL, $c$. Our primary Cancel-VL metric evaluates only the seven fact labels; $c$ remains a required auxiliary output and is checked separately. The main challenges are cross-modal evidence grounding, fine-grained rule discrimination, and label imbalance.

Map tasks emphasize \emph{reasoning depth} through sequential route dependencies and strict whole-route verification, whereas VL tasks emphasize \emph{rule breadth} through multiple decisions coupled by shared business rules. \method{} fixes inputs, output contracts, and verifiers while alternating updates to the external program $\Phi_d$ and policy $\theta$. Let $\mathcal D_{\mathrm{mm}}=\mathcal D_d^{\mathrm{train}}$ denote the training pool, and freeze $\mathcal D_{\mathrm{val}}$ before candidate generation. Test data never enter Harness construction, parameter training, or candidate selection. Optimization therefore alternates discrete Harness commits with parameter updates rather than jointly differentiating $\theta$ and $\Phi_d$.

\subsection{Overview: Build--Learn--Reconstruct--Learn Again}

\begin{figure}[ht]
\centering
\includegraphics[width=0.99\textwidth]{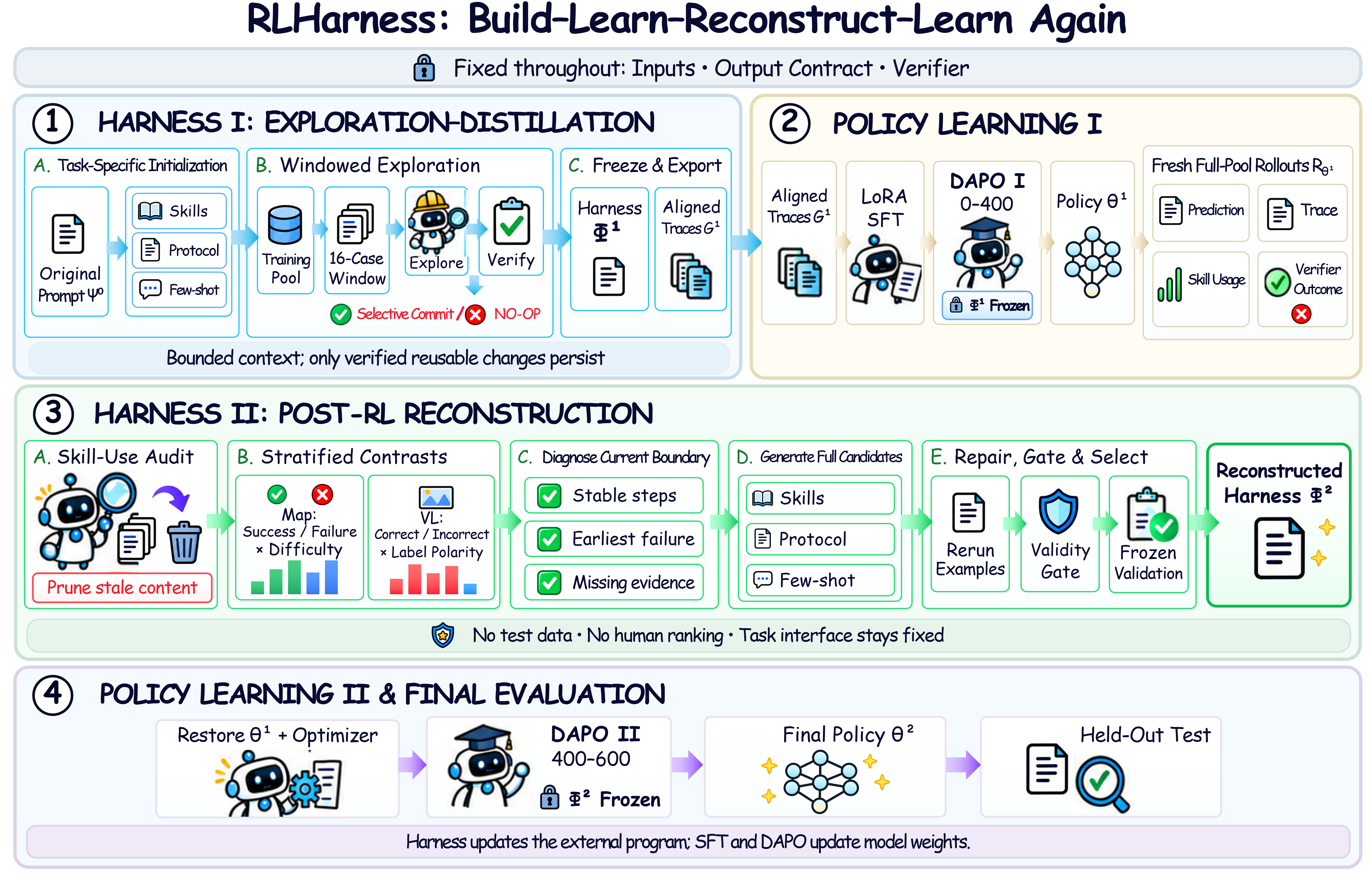}
\setlength{\abovecaptionskip}{3pt}
\setlength{\belowcaptionskip}{0pt}
\caption{The complete \method{} pipeline. \textcircled{1} \textbf{Harness I} initializes the task-specific external program, fixes a task-specific exploration subset and batch plan, and selectively commits validated changes. After the first Harness $\Phi^1$ is frozen, audited accepted traces $\mathcal G^1$ are exported from complete-training-pool attempts. \textcircled{2} \textbf{Policy Learning I} distills these traces and runs DAPO I with $\Phi^1$ frozen, producing policy $\theta^1$ and fresh rollouts. \textcircled{3} \textbf{Harness II} audits Skill use, contrasts stratified successes and failures, diagnoses the current capability boundary, and selects a validated reconstructed Harness $\Phi^2$. \textcircled{4} \textbf{Policy Learning II} resumes from $\theta^1$ and runs DAPO II with $\Phi^2$ frozen. Only the resulting $\theta^2$--$\Phi^2$ pair is evaluated on the held-out test set. Inputs, output contracts, and verifiers remain fixed throughout.}
\label{fig:rlharness-pipeline}
\end{figure}

\method{} maintains both an external program and model parameters. The \emph{external Harness} organizes task rules, invocation methods, and examples into a program that the model can execute. The \emph{parametric policy} learns how to execute that program and, after each update, produces new behavior for the next Harness reconstruction. For domain $d$, we write
\[
\mathcal H_d^j=(\Phi_d^j,\Gamma_d^j),\qquad
\Phi_d^j=(I_d,B_d^j,P_d^j,E_d^j,C_d).
\]
Here $B$ is the Skill Bank, $P$ is the Skill-selection and execution protocol, and $E$ contains few-shot examples that demonstrate concrete invocations; $I$ and $C$ fix input assembly and the output contract, respectively. Control plane $\Gamma$ organizes analysis contexts, calls the verifier, repairs or rolls back failed revisions, and commits complete versions atomically. The two Harness stages change only the external state $B/P/E$, whereas SFT and DAPO change only the model parameters $\theta$.

The complete lifecycle is
\[
(\mathcal D_{\mathrm{mm}},\Psi^0)
\xrightarrow{\mathcal H_{\mathrm{ED}}}(\Phi^1,\mathcal G^1)
\xrightarrow{\mathrm{SFT+DAPO\ I}}(\theta^1,\mathcal R_{\theta^1})
\xrightarrow{\mathcal H_{\mathrm{PR}}}\Phi^2
\xrightarrow{\mathrm{DAPO\ II}}\theta^2.
\]
This chain has four steps. Harness I first turns raw task experience into an initial program describing \emph{how the model should solve the task}. Policy Learning I internalizes that program and exposes new errors. Harness II then rebuilds the program according to what the current policy has already mastered and where it still fails. Policy Learning II finally adapts the model to the revised program.

The two Harness stages are therefore not repeated runs of the same Prompt self-optimizer. Harness I emphasizes \emph{task coverage} and builds a complete program for an unadapted model; Harness II emphasizes the \emph{current policy's capability boundary} and repairs only the mismatches that remain or emerge after the first learning block.

\subsection{Harness I: Building a Learnable Program from Raw Experience}

Harness I addresses program cold start before parameter training. Rather than training the model directly, it first turns the original Prompt and training experience into a first Harness that can be executed, verified, and imitated, and it generates supervision under exactly the same version:
\[
\mathcal H_{\mathrm{ED}}:
(\mathcal D_{\mathrm{mm}},\mathcal D_{\mathrm{val}},\Psi^0;S,T,V)
\longmapsto(\Phi^1,\mathcal G^1).
\]
Here $S/T/V$ denote the exploration executor, task teacher, and verifier. In $\Phi^1=(I_d,B_d^1,P_d^1,E_d^1,C_d)$, input assembly $I_d$ and output contract $C_d$ remain fixed; Harness I constructs the \textbf{Skill Bank} $B_d^1$, \textbf{invocation protocol} $P_d^1$, and \textbf{few-shot block} $E_d^1$.

Harness I first creates and verifies the initial $B/P/E$ package, then explores a fixed task-specific subset in preassigned batches. For the released map pipeline, seed 42 selects 480 Train1600 examples proportionally by difficulty and orders them into twelve batches of 40. The Prompt is fixed within each batch; the explorer executes the batch, the teacher reflects on verified outcomes, and a complete Prompt candidate is produced by reflection, merging, and ranking. A candidate must preserve the immutable task interface and is committed only when its hard accuracy on the complete frozen Val100 is strictly higher than the incumbent; a no-op, tie, decrease, or invalid candidate leaves the Prompt unchanged. Batch-local trajectories are then evicted. This released path creates and verifies the two fixed demonstrations during initialization; it does not rerun them after every accepted SkillOpt edit. Once $\Phi^1$ is frozen, the teacher attempts the complete training pool under exactly that version and retains the audited accepted trajectories as $\mathcal G^1$ for SFT. The private VL pipeline uses the same state boundary with its task-specific batching and verifier.

Harness I thus forms a reusable program under bounded context and passes it to SFT and DAPO I together with examples and trajectories from the same version. Appendix~\ref{app:harness-one-details} details the batch plan, commit gate, and trajectory audit; Appendix~\ref{app:stage-algorithms} gives the stage pseudocode.

\subsection{Harness II: Reconstructing the Learning Environment from Current Behavior}

Harness II addresses the version mismatch after the first learning block. Once SFT and DAPO I produce $\theta^1$, the policy has internalized part of $\Phi^1$, and the remaining error distribution has changed. Harness II therefore does not reuse the stale pre-training failures; it reconstructs the Harness required by the next RL block from fresh rollouts of the current policy:
\[
\mathcal H_{\mathrm{PR}}:
(\mathcal R_{\theta^1},\mathcal D_{\mathrm{mm}},\Phi^1,
\mathcal D_{\mathrm{val}};T,Q,V)
\longmapsto\Phi^2.
\]
Input assembly, the output contract, source examples, and the verifier remain fixed. Only the \textbf{Skill Bank}, \textbf{invocation protocol}, and \textbf{few-shot block} are reconstructed.

Harness II follows the order of identifying the current boundary, repairing the external program, and helping the policy switch. It first performs \textbf{stratified sampling and summarization}: comparable successes and failures from the current rollouts are contrasted batch by batch to record which steps are stable, where execution still fails, and what evidence is missing or misused. It then performs \textbf{Skill correction}: ineffective Skills are pruned using both the summaries and observed usage, while triggers, execution steps, and invocation order are revised without disturbing procedures that already work. Finally, \textbf{version-matched few-shot demonstrations} help the policy switch: each candidate receives a separately rewritten How-to block and demonstration section whose markers and local reasoning follow the new Skill boundaries and execution order. Full parameter adaptation remains the responsibility of DAPO II.

Only candidates that pass the deterministic candidate and few-shot gates enter the frozen validation pool. Qwen3.5-plus executes each valid candidate and produces a normalized report; a separate task-teacher call---GPT-5.6-sol in the released map pipeline---selects from the complete reports and Prompts. The checkpoint-400 student does not execute candidate validation; it supplies reconstruction evidence and later consumes the selected $\Phi^2$ during DAPO II. Neither the test set nor human ranking participates in selection. Harness II is therefore a pipeline of stratified diagnosis, joint Skill/protocol correction, same-version examples, proxy execution, and automatic commitment. Appendix~\ref{app:harness-two-details} describes candidate repair and validation, and Appendix~\ref{app:harness-evidence-controls} gives the task-specific sampling budgets.

\subsection{Policy Learning: Converting Both Harnesses into Model Capability}

Policy learning connects the two Harness versions. It trains the model to follow the external program and later provides fresh behavior for reconstruction. The process consists of reasoning distillation, a shared reward and Hybrid-DGPO objective, and two consecutive learning blocks.

\paragraph{Reasoning distillation.} An external pipeline audits $\mathcal G^1$ and converts it into SFT data. The targets cover Skill selection, multimodal evidence binding, intermediate computation, and answer checking, so the policy learns to execute $\Phi^1$ rather than copy final answers alone.

\paragraph{Shared task reward and Hybrid-DGPO.} Both DAPO blocks use
    \[
    R=0.05r_{\mathrm{format}}+0.25r_{\mathrm{partial}}+0.70r_{\mathrm{all}}.
    \]
    Map scorers verify partial and exact routes; VL scorers verify the fact vector. For every task, question $q$ receives rollout difficulty $d_q=1-p_q$ and
    \[
    \widehat A_{qi}=w_q\bigl(R_{qi}-\bar R_q\bigr).
    \]
    Fee-VL and Cancel-VL additionally apply per-label frequency weights inside $r_{\mathrm{partial}}$, before the scalar reward is centered; these label weights are not a replacement for $w_q$. Thus all four tasks use difficulty-weighted advantages, while only the two VL tasks also rebalance rare positive fact labels. Appendix~\ref{app:dapo-config} gives the per-task definitions and the exact ablation.

\paragraph{Two parameter-learning blocks.} The first block runs SFT on $\mathcal G^1$, then keeps $\Phi^1$ fixed during DAPO I to obtain $\theta^1$ and fresh rollouts. The second keeps $\Phi^2$ fixed and resumes from the checkpoint-400 model and optimizer state:
    \[
    \theta^1=\operatorname{DAPO}_{0:400}(\theta_{\mathrm{SFT}};\Phi^1),
    \qquad
    \theta^2=\operatorname{DAPO}_{400:600}(\theta^1;\Phi^2).
    \]

The feedback remains asymmetric: the Harness defines the program used in the next learning block, and policy rollouts reveal what the next Harness should repair. Parameter optimization remains with the policy, while final deployment and evaluation use the frozen $\theta^2$--$\Phi^2$ pair. Appendix~\ref{app:end-to-end} gives the schedule; Appendices~\ref{app:harness-two-details} and~\ref{app:harness-evidence-controls} describe candidate construction, checks, and evidence budgets; Appendix~\ref{app:stage-algorithms} \mbox{gives pseudocode}.

\section{Experiments}

\subsection{Experimental Setup}

\paragraph{Research questions.}
We organize the experiments around four questions. \textbf{RQ1:} Does the complete build--learn--reconstruct--learn path improve performance, and which components are necessary? \textbf{RQ2:} How does the final system compare with representative methods and models? \textbf{RQ3:} What capability transfers through the learned weights? \textbf{RQ4:} How task-specific is the reconstructed Post-RL Prompt? The following subsections present the corresponding evidence; Appendices~\ref{app:model-transfer} and~\ref{app:skill-transfer} provide the full transfer matrices.

\paragraph{Benchmarks.}
We evaluate four long-horizon cross-modal reasoning benchmarks spanning public map planning and real ride-hailing operations:
\begin{enumerate}
\item \textbf{MetroMap-lite} is a sampled subset of MetroMap, a multimodal benchmark for metro-diagram understanding and weighted route planning; we retain the \texttt{-lite} suffix throughout.
\item \textbf{TravelMap-lite} is a sampled subset of TravelMap, a multimodal benchmark for travel-map understanding and route planning; we retain the \texttt{-lite} suffix throughout.
\item \textbf{Fee-VL} is a multimodal business dataset formed from Didi Chuxing's handling of fare disputes between drivers and passengers.
\item \textbf{Cancel-VL} is a multimodal business dataset constructed for responsibility assignment after driver--passenger order cancellations at Didi Chuxing.
\end{enumerate}
Appendix~\ref{app:data-statistics} gives the complete data distributions, label composition, and difficulty stratification.

\paragraph{Training and evaluation.}
Every trainable branch uses two epochs of LoRA SFT, the same DAPO reward, eight samples per question, and Hybrid-DGPO. DAPO I ends at step 400; DAPO II resumes from the same model and optimizer state and continues to step 600. MetroMap-lite and TravelMap-lite report strict whole-route accuracy. Fee-VL and Cancel-VL report Overall Macro F1, the unweighted mean of the fact-label F1 scores. Cancel-VL responsibility classification is a required auxiliary output for resolved samples and is checked by the verifier, but it is not included in the reported primary metric. Each training prompt contains exactly one image. Test data are excluded from Harness updates, model training, and candidate selection. Appendices~\ref{app:sft-config} and~\ref{app:dapo-config} list the SFT and DAPO hyperparameters; Appendices~\ref{app:harness-one-details} and~\ref{app:harness-two-details} describe Harness model roles and candidate selection; Appendix~\ref{app:statistical-tests} reports the paired statistical tests.

\subsection{RQ1: Does the Build--Learn--Reconstruct--Learn Path Work?}

Table~\ref{tab:joint-ablation} reports the main lifecycle and matched ablations in build--learn--reconstruct--learn order. ED (Harness I) tests program cold start and the policy's internalization of $\Phi^1$. PR (Harness II) reconstructs the program from current-policy rollouts, and DAPO II then tests adaptation to $\Phi^2$. Each minus row removes only the named component, superscripts report gains over Base, and Appendix~\ref{app:matched-ablations} specifies the complete controls. Appendix~\ref{app:statistical-tests} reports $K=10$ matched runs: all 24 primary component comparisons are significant after Holm correction.

\paragraph{Overall results.}
The two task families have different bottlenecks. Map planning depends on a stable chain of graph recovery, constraint checking, and route search; partial-route and whole-route rewards can correct errors along that chain, so the two RL blocks contribute most strongly. VL must execute complex business rules over image and order evidence; RL in a stale environment cannot repair Skill coverage or decision boundaries, so the Skills, protocol, and few-shot block must first be reconstructed and then internalized by RL. Under the same budget, $\Phi^2$ also outperforms continued use of $\Phi^1$, showing that the gain is not explained by 200 extra steps alone.

\begin{table*}[t]
\centering
\footnotesize
\renewcommand{\arraystretch}{1.04}
\setlength{\tabcolsep}{3pt}
\begin{tabularx}{\textwidth}{>{\raggedright\arraybackslash}Xcccc}
\toprule
\rowcolor{tableheadergray}
\makecell[l]{\textbf{Stage /}\\\textbf{ablation}}
& \makecell[c]{\textbf{MetroMap-lite}\\\textbf{(Acc/\%)}}
& \makecell[c]{\textbf{TravelMap-lite}\\\textbf{(Acc/\%)}}
& \makecell[c]{\textbf{Fee-VL}\\\textbf{(F1-Score/\%)}}
& \makecell[c]{\textbf{Cancel-VL}\\\textbf{(F1-Score/\%)}} \\
\midrule
Base & 16.25 & 27.50 & 37.13 & 45.50 \\
\hspace{0.7em}$+$ ED Init. & 24.50\vsbase{8.25} & 30.50\vsbase{3.00} & 38.83\vsbase{1.70} & 46.99\vsbase{1.49} \\
\hspace{0.7em}$+$ ED Explore & 31.50\vsbase{15.25} & 33.25\vsbase{5.75} & 40.43\vsbase{3.30} & 49.00\vsbase{3.50} \\
\hspace{1.4em}$-$ Selective Commit & 29.75\vsbase{13.50} & 32.25\vsbase{4.75} & 39.25\vsbase{2.12} & 47.75\vsbase{2.25} \\
\lightrowrule
$+$ Distillation & 37.50\vsbase{21.25} & 35.50\vsbase{8.00} & 47.43\vsbase{10.30} & 58.30\vsbase{12.80} \\
\hspace{0.7em}$-$ Skill-guided Trace & 29.75\vsbase{13.50} & 31.25\vsbase{3.75} & 42.57\vsbase{5.44} & 53.12\vsbase{7.62} \\
\lightrowrule
$+$ DAPO I & 57.50\vsbase{41.25} & 43.00\vsbase{15.50} & 49.23\vsbase{12.10} & 59.49\vsbase{13.99} \\
\hspace{0.7em}$-$ Question-Difficulty Weighting & 53.25\vsbase{37.00} & 40.75\vsbase{13.25} & 46.57\vsbase{9.44} & 56.74\vsbase{11.24} \\
\lightrowrule
$+$ PR Rebuild & 55.25\vsbase{39.00} & 42.75\vsbase{15.25} & 48.31\vsbase{11.18} & 60.21\vsbase{14.71} \\
\hspace{0.7em}$-$ Stratified Contrastive Evidence & 54.25\vsbase{38.00} & 42.00\vsbase{14.50} & 47.50\vsbase{10.37} & 59.00\vsbase{13.50} \\
\hspace{0.7em}$-$ Few-shot & 52.50\vsbase{36.25} & 40.50\vsbase{13.00} & 46.86\vsbase{9.73} & 56.92\vsbase{11.42} \\
\lightrowrule
$+$ DAPO II & \textbf{62.00\vsbase{45.75}} & \textbf{50.00\vsbase{22.50}} & \textbf{65.81\vsbase{28.68}} & \textbf{65.51\vsbase{20.01}} \\
\hspace{0.7em}Random-Evidence Prompt + DAPO II & 60.50\vsbase{44.25} & 48.75\vsbase{21.25} & 63.50\vsbase{26.37} & 63.75\vsbase{18.25} \\
\hspace{0.7em}Budget control: Reuse $\Phi^1$ & 59.00\vsbase{42.75} & 47.50\vsbase{20.00} & 51.06\vsbase{13.93} & 60.37\vsbase{14.87} \\
\bottomrule
\end{tabularx}
\caption{RLHarness lifecycle and matched ablations. Superscripts show absolute gains (\%) over Base. Minus rows denote the five component ablations. Random-Evidence Prompt + DAPO II is the continuation branch of the stratified contrastive-evidence ablation; the training control is not counted as an ablation. Acc is strict whole-route accuracy and F1-Score is fact-label macro F1.}
\label{tab:joint-ablation}
\end{table*}

\paragraph{ED initialization, ED exploration, and selective commitment.}
ED initialization turns the raw Prompt into an executable Skill Bank, protocol, How-to block, and verified demonstrations. ED exploration then processes a fixed 480-example subset under a task-specific batch plan; the released map runs use twelve batches of 40 and commit a valid candidate only after a strict Val100 hard-accuracy gain. The matched ablation preserves the evidence and calls but disables rejection of a legal effective edit. Exploration improves on Base by an average of 6.95\%, and selective commitment adds 1.30\%, showing that retaining the incumbent when an edit lacks support is useful.

\paragraph{Reasoning distillation and Skill-guided trajectories.}
Reasoning distillation uses audited solutions produced under $\Phi^1$ as SFT targets. These Skill-guided trajectories include the answer, the chosen Skills, the order in which evidence is inspected, and the protocol used to reach the result. In the ablation, the same teacher solves each example freely, without following the Harness. Version-aligned trajectories improve the average score by 5.51\%, showing that SFT must teach the deployed procedure, not just a correct answer.

\paragraph{DAPO I and Hybrid-DGPO.}
DAPO I is the first RL block and keeps $\Phi^1$ fixed through step 400. Hybrid-DGPO gives more weight to questions the current policy often misses; the ablation uses the same unnormalized within-question estimator with $w_q=1$. DAPO I improves the map tasks by 13.75\% on average but the VL tasks by only 1.50\%; Question-difficulty weighting adds 2.98\% over this matched control. For maps, partial- and exact-route rewards give useful feedback before the full chain is correct. On VL, imbalanced positive labels weaken the within-group comparison. Difficulty weighting directs training toward unresolved questions, but cannot fix a bad Skill boundary in $\Phi^1$.

\paragraph{PR reconstruction and immediate ablations.}
After DAPO I, PR reconstruction rewrites the Skills, protocols, and demonstrations from the policy's latest successes and failures. For maps, the evidence pairs successes and failures across difficulty levels. For VL, it balances correct and incorrect predictions and, where possible, positive and negative labels. The rebuilt few-shot examples show when the new rules apply and in what order to execute them. Swapping $\Phi^1$ for $\Phi^2$ without updating $\theta^1$ lowers the average score by 0.68\%, which exposes the version mismatch directly. Even before retraining, however, stratified evidence is 0.94\% better than random evidence, and retaining the synchronized demonstrations is 2.44\% better than removing them.

\paragraph{DAPO II: adapting to the reconstructed Harness.}
DAPO II trains the policy on the reconstructed Harness, continuing from checkpoint 400 to 600 with $\Phi^2$ fixed. The random-evidence branch uses the same schedule but builds its Prompt from unstratified examples. The budget control also adds 200 steps, but keeps $\Phi^1$. Relative to the immediate Harness swap, DAPO II raises the map and VL averages by 7.00\% and 11.40\%. The full method is also 1.71\% above the random-evidence branch and exceeds reuse of $\Phi^1$ by 2.75\% on maps and 9.95\% on VL, including 14.75\% on Fee-VL. The policy must therefore learn the reconstructed program; extra steps under the old Prompt are not~enough.

\subsection{RQ2: Comparison with Representative Methods and Models}
\label{sec:model-comparison}

\begin{table*}[t]
\centering
\footnotesize
\renewcommand{\arraystretch}{1.04}
\setlength{\tabcolsep}{2.4pt}
\begin{minipage}[t]{0.485\textwidth}
\centering
\textbf{(a) Long-horizon reasoning tasks}\\[3pt]
\begin{tabularx}{\linewidth}{>{\raggedright\arraybackslash}Xcc}
\toprule
\rowcolor{tableheadergray}
\makecell[c]{\textbf{Model}} & \makecell[c]{\textbf{Metro-Lite}\\\textbf{(Acc/\%)}} & \makecell[c]{\textbf{Travel-Lite}\\\textbf{(Acc/\%)}} \\
\midrule
GPT-4o & 20.50 & 17.00 \\
GPT-4.1 & 27.50 & 25.75 \\
Qwen3-Max & 27.75 & 38.50 \\
Gemini-3-Flash-Preview & 74.25 & 52.50 \\
GPT-5.5 & \textbf{78.00} & \textbf{64.75} \\
\lightrowrule
SkillRL$^\dagger$ & 24.59 & 36.67 \\
\textbf{\method{} (Ours)} & 62.00 & 50.00 \\
\bottomrule
\end{tabularx}
\end{minipage}
\hfill
\begin{minipage}[t]{0.485\textwidth}
\centering
\textbf{(b) Multi-scenario reasoning tasks}\\[3pt]
\begin{tabularx}{\linewidth}{>{\raggedright\arraybackslash}Xcc}
\toprule
\rowcolor{tableheadergray}
\makecell[c]{\textbf{Model}} & \makecell[c]{\textbf{Fee-VL}\\\textbf{(F1-Sco./\%)}} & \makecell[c]{\textbf{Cancel-VL}\\\textbf{(F1-Sco./\%)}} \\
\midrule
Qwen3.6-35B-A3B & 46.60 & 52.84 \\
Kimi-K2.5 & 49.36 & 58.47 \\
Qwen3.8-27B & 54.72 & 63.26 \\
Kimi-K3 & 59.18 & 67.41 \\
Qwen3.8-Max-0902 & 61.47 & \textbf{69.18} \\
\lightrowrule
SkillRL$^\dagger$ & 42.81 & 51.92 \\
\textbf{\method{} (Ours)} & \textbf{65.81} & 65.51 \\
\bottomrule
\end{tabularx}
\end{minipage}
\caption{Overall performance on four reasoning tasks. Metro-Lite and Travel-Lite denote MetroMap-lite and TravelMap-lite. $\dagger$ marks SkillRL with its original implementation and configuration, changing only the task datasets and interfaces; bold marks the best result in each column.}
\label{tab:model-comparison}
\end{table*}

Table~\ref{tab:model-comparison} contrasts the final Qwen3.5-9B system with SkillRL and representative general models. Relative to SkillRL, RLHarness improves Metro-Lite/Travel-Lite by 37.41/13.33 Accuracy points and Fee-VL/Cancel-VL by 23.00/13.59 Macro-F1 points. SkillRL retrieves and updates individual Skills, whereas RLHarness jointly reconstructs Skills, protocols, and demonstrations from current-policy traces and then adapts the policy to the revised program. The station- and edge-level or label-boundary precision required by these tasks exposes the limitation of individual-Skill retrieval and revision, while the results support version-aligned program reconstruction followed by policy adaptation. The general-model rows only indicate how closely the adapted 9B open-weight model approaches strong-model capability; they are not matched method baselines. Evaluation and baseline-scope details are provided in Appendices~\ref{app:external-reference-audit} and~\ref{app:limitations}.

\subsection{RQ3 and RQ4: Separate Evaluation of Model-Weight and Prompt Transfer}

For RQ3, we transfer checkpoint-600 weights while restoring the initial Prompt of the target task, isolating parameter transfer from Prompt transfer. The weights improve every tested off-diagonal target over the unadapted baseline, but the gains are modest and strongest within the same task family; Table~\ref{tab:model-transfer} in Appendix~\ref{app:model-transfer} gives the full matrix. For RQ4, we fix the target checkpoint, data, reward, and scorer, replace only the Post-RL Prompt source, and run the matched checkpoint-400-to-600 continuation. The task-matched Prompt wins in all four within-family replacements, with smaller losses between the two map tasks than between Fee-VL and Cancel-VL; Table~\ref{tab:skill-transfer} in Appendix~\ref{app:skill-transfer} gives the controlled comparison. Together, the results show that learned weights carry limited, task-family-biased reusable capability, whereas the reconstructed Prompt remains strongly task-specific; Appendices~\ref{app:detailed-method} and~\ref{app:discussion} provide implementation details and limitations.

\section{Conclusion}

\method{} closes the loop between external-program maintenance and parameter learning. Harness I builds a version-aligned program and supervision from raw experience; Harness II reconstructs it from current-policy successes and failures. SFT and two DAPO blocks internalize both versions. Across all four tasks, all 24 primary component comparisons are significant in $K=10$ paired runs after Holm correction, and every task peaks only after reconstruction and DAPO II; reusing the old Harness is weaker under the same budget. Transfer results distinguish the two capability stores: learned weights transfer positively but mainly within task families, whereas Post-RL Prompts transfer poorly across different rules and label boundaries. Thus, gains come from versioned Harness--policy adaptation rather than a one-time Prompt rewrite or longer training. Limitations include visual-reading errors, coarse terminal feedback, and only two update cycles. Future work should add explicit visual representations, stage-level verifiers, and longer auditable lifecycles.

\clearpage
\nocite{li2026skillsbench,han2026sweskillsbench,zhang2026evoskills,li2026skillgraph,li2026arise,vishe2026skillr1,yang2026skillforge,huang2026skillsp,fu2026sesa,shen2026skillalpha,zhang2026skillflow,huang2026memoharness,lin2026agenticharness,liu2026adaptiveharness,wei2026evoharness,yi2026offlineharness,paul2026controlsystem,zhang2026flowsteer,ning2026evoharnessrl}
\bibliography{skill_model_evo_paper_latex}
\bibliographystyle{iclr2027_conference}

\clearpage
\appendix
\onecolumn

\section*{\centering \LARGE Appendix Contents}
\addcontentsline{toc}{section}{Appendix Contents}

\vspace{0.45cm}
\begingroup
\setlength{\parindent}{0pt}
\setlength{\parskip}{0.10em}
\fontsize{10.5pt}{14pt}\selectfont

\noindent\hyperref[app:reproducibility]{\textbf{A \quad Reproducibility Statement}}\dotfill\pageref{app:reproducibility}\par

\vspace{0.55em}

\noindent\hyperref[app:ai-use]{\textbf{B \quad AI Use Statement}}\dotfill\pageref{app:ai-use}\par

\vspace{0.55em}

\noindent\hyperref[app:unified-config]{\textbf{C \quad Experimental Setup and Additional Results}}\dotfill\pageref{app:unified-config}\par
\noindent\hspace*{1.5em}\hyperref[app:data-statistics]{C.1. \quad Dataset and Label Statistics}\dotfill\pageref{app:data-statistics}\par
\noindent\hspace*{1.5em}\hyperref[app:fee-label-results]{C.2. \quad Fee-VL Label-Level Results}\dotfill\pageref{app:fee-label-results}\par
\noindent\hspace*{1.5em}\hyperref[app:cancel-label-results]{C.3. \quad Cancel-VL Label-Level Results}\dotfill\pageref{app:cancel-label-results}\par
\noindent\hspace*{1.5em}\hyperref[app:skill-transfer]{C.4. \quad Cross-Task Post-RL Prompt Replacement}\dotfill\pageref{app:skill-transfer}\par
\noindent\hspace*{1.5em}\hyperref[app:model-transfer]{C.5. \quad Cross-Task Evaluation of Trained Model Weights}\dotfill\pageref{app:model-transfer}\par
\noindent\hspace*{1.5em}\hyperref[app:sft-config]{C.6. \quad SFT Configuration}\dotfill\pageref{app:sft-config}\par
\noindent\hspace*{1.5em}\hyperref[app:dapo-config]{C.7. \quad DAPO Configuration}\dotfill\pageref{app:dapo-config}\par
\noindent\hspace*{1.5em}\hyperref[app:stage-mapping]{C.8. \quad Evaluation Protocol and Stage Mapping}\dotfill\pageref{app:stage-mapping}\par
\noindent\hspace*{1.5em}\hyperref[app:matched-ablations]{C.9. \quad Matched Ablation Settings}\dotfill\pageref{app:matched-ablations}\par
\noindent\hspace*{1.5em}\hyperref[app:statistical-tests]{C.10. \quad Paired Statistical Tests}\dotfill\pageref{app:statistical-tests}\par

\vspace{0.55em}

\noindent\hyperref[app:detailed-method]{\textbf{D \quad Method Details}}\dotfill\pageref{app:detailed-method}\par
\noindent\hspace*{1.5em}\hyperref[app:end-to-end]{D.1. \quad End-to-End Training Algorithm}\dotfill\pageref{app:end-to-end}\par
\noindent\hspace*{1.5em}\hyperref[app:harness-boundary]{D.2. \quad State Representation and Commit Protocol}\dotfill\pageref{app:harness-boundary}\par
\noindent\hspace*{1.5em}\hyperref[app:harness-one-details]{D.3. \quad Exploration--Distillation Harness}\dotfill\pageref{app:harness-one-details}\par
\noindent\hspace*{1.5em}\hyperref[app:policy-learning-one]{D.4. \quad First-Stage Policy Learning}\dotfill\pageref{app:policy-learning-one}\par
\noindent\hspace*{1.5em}\hyperref[app:harness-two-details]{D.5. \quad Post-RL Harness Reconstruction}\dotfill\pageref{app:harness-two-details}\par
\noindent\hspace*{1.5em}\hyperref[app:policy-learning-two]{D.6. \quad Second-Stage Policy Learning}\dotfill\pageref{app:policy-learning-two}\par
\noindent\hspace*{1.5em}\hyperref[app:harness-evidence-controls]{D.7. \quad Evidence Budgets and Matched Controls}\dotfill\pageref{app:harness-evidence-controls}\par
\noindent\hspace*{1.5em}\hyperref[app:stage-algorithms]{D.8. \quad Stage-Specific Algorithms}\dotfill\pageref{app:stage-algorithms}\par

\vspace{0.55em}

\noindent\hyperref[app:prompt-cases]{\textbf{E \quad Prompt Specification and Case Studies}}\dotfill\pageref{app:prompt-cases}\par
\noindent\hspace*{1.5em}\hyperref[app:metromap-prompt]{E.1. \quad MetroMap-lite Prompt Specification}\dotfill\pageref{app:metromap-prompt}\par
\noindent\hspace*{1.5em}\hyperref[app:end-to-end-cases]{E.2. \quad Case Studies}\dotfill\pageref{app:end-to-end-cases}\par

\vspace{0.55em}

\noindent\hyperref[app:discussion]{\textbf{F \quad Discussion, Limitations, and Future Directions}}\dotfill\pageref{app:discussion}\par
\noindent\hspace*{1.5em}\hyperref[app:interpretation]{F.1. \quad Discussion}\dotfill\pageref{app:interpretation}\par
\noindent\hspace*{1.5em}\hyperref[app:limitations]{F.2. \quad Limitations}\dotfill\pageref{app:limitations}\par
\noindent\hspace*{1.5em}\hyperref[app:future-work]{F.3. \quad Future Directions}\dotfill\pageref{app:future-work}\par

\vspace{0.55em}

\noindent\hyperref[app:acknowledgments]{\textbf{G \quad Acknowledgments}}\dotfill\pageref{app:acknowledgments}\par
\endgroup
\vspace{0.8cm}
\clearpage

\section{Reproducibility Statement}
\label{app:reproducibility}

The supplementary material specifies the data distributions, Harness state and stage boundaries, evidence budgets, candidate-validation procedure, SFT and DAPO configurations, checkpoint schedule, rewards, ablations, and task scorers used in the reported experiments. The public release provides the end-to-end MetroMap-lite and TravelMap-lite implementation, including configurations, Prompt and Harness snapshots, split identifiers, evaluation scripts, and auditable run artifacts. The Fee-VL and Cancel-VL data and private business pipeline cannot be released under privacy and governance constraints; for these tasks, we provide aggregate statistics, de-identified examples, and the processing and evaluation protocols used in this study.

\section{AI Use Statement}
\label{app:ai-use}

Generative AI tools assisted with language editing, code-level consistency checks, literature search, and LaTeX preparation. The authors designed the method and experiments, verified technical claims and numerical values against the authoritative experiment records, selected the cited literature, and take responsibility for the manuscript.

\section{Experimental Setup and Additional Results}
\label{app:unified-config}

\subsection{Dataset and Label Statistics}
\label{app:data-statistics}

\subsubsection{Map-Task Dataset Statistics}

MetroMap-lite and TravelMap-lite define difficulty as the Cartesian product of map difficulty and query difficulty. The lite construction primarily excludes extreme cases whose input context or required reasoning chain is too long. Under the common length budget, such cases are prone to truncation, so their outcomes reflect context and generation limits more than the route-reasoning capability studied here. The filter does not depend on whether the model answers correctly and does not specifically remove easy cases. The remaining cases are sampled by difficulty stratum to form the fixed Train1600 and Test400 splits below. Accordingly, all map conclusions in this paper apply to the retained easy/medium/hard strata under the stated context and response limits; they do not establish performance on the excluded extreme-length portion of the full benchmarks.

For MetroMap-lite, all 1,600 Train1600 prompts enter DAPO and 1,395 audited traces enter SFT. Harness-I exploration uses seed 42 to sample and fix 480 examples from Train1600 proportionally by difficulty; this subset does not change the data used by DAPO or SFT. Val100 is drawn from the original training source, is sample-ID disjoint from the locked Train1600 and Test400, and is frozen before optimization. It is used both for the Harness-I strict commit gate and for post-RL Prompt selection. TravelMap-lite additionally isolates figures across the split, whereas the MetroMap-lite Train480 and Val100 share 58 figures. For TravelMap-lite, 1,233 audited traces enter SFT; the task uses the same Train1600/Val100/Test400 split roles and the same 480-example Harness-I exploration budget. Both Test400 sets are used only for final evaluation. Because the same Val100 supports decisions in both Harness stages, it is a development set rather than an independent performance estimate; final claims use Test400.

\begin{table}[ht]
\centering
\footnotesize
\renewcommand{\arraystretch}{1.10}
\setlength{\tabcolsep}{4.0pt}
\begin{tabularx}{0.82\textwidth}{L*{4}{C}}
\toprule
\multicolumn{1}{c}{\multirow{2}{*}{\textbf{Stratum}}}
& \multicolumn{2}{>{\columncolor{tableheadergray}}c}{\textbf{MetroMap-lite}}
& \multicolumn{2}{>{\columncolor{tableheadergray}}c}{\textbf{TravelMap-lite}} \\
\cmidrule(lr){2-3}\cmidrule(lr){4-5}
& \cellcolor{tableheadergray}\textbf{Train}
& \cellcolor{tableheadergray}\textbf{Test}
& \cellcolor{tableheadergray}\textbf{Train}
& \cellcolor{tableheadergray}\textbf{Test} \\
\midrule
E$\times$E & 455 & 133 & 471 & 110 \\
\lightrowrule
E$\times$M & 321 & 74 & 325 & 90 \\
\lightrowrule
E$\times$H & 97 & 17 & 50 & 16 \\
\lightrowrule
M$\times$E & 292 & 63 & 301 & 76 \\
\lightrowrule
M$\times$M & 313 & 84 & 380 & 78 \\
\lightrowrule
H$\times$E & 122 & 29 & 73 & 30 \\
\midrule
\textbf{Total} & \textbf{1,600} & \textbf{400} & \textbf{1,600} & \textbf{400} \\
\bottomrule
\end{tabularx}
\caption{Difficulty distributions of the MetroMap-lite and TravelMap-lite splits. E/M/H denote easy/medium/hard, and $\times$ separates map and query difficulty; Train and Test denote the fixed Train1600 and Test400 splits.}
\label{tab:map-lite-data}
\end{table}

\subsubsection{Vision--Language Dataset Statistics}

Fee-VL and Cancel-VL both use ride-hailing orders as their basic examples, jointly presenting trajectory, report, or map images with structured fields such as time, distance, and order status. Neither task is ordinary image classification: the model must integrate cross-modal evidence under platform rules and make a separate decision for each predefined label. Because one order may activate several fact labels, we report both the number of orders/images and the number of order--label decisions; the latter counts the independent binary judgments actually evaluated across labels.

\paragraph{Prompt construction and data roles.}
The mathematical input notation denotes the visual field available to a task, but the implemented benchmark uses exactly one image per prompt. A frozen preprocessing record selects the referenced trajectory, map, or report image for that prompt; images are not concatenated and no prompt supplies a multi-image list. Structured fields and labels remain attached to the same order record.

The 1,600 RL prompts do not denote the complete VL corpus. They are fixed task-specific subsets of the complete training pools, selected and frozen before optimization; the released run uses their fixed ID lists for both DAPO blocks. Harness-I exploration fixes 480 prompts from each RL set, whereas SFT uses the separately audited trace corpus. Harness validation is frozen before Prompt optimization; for maps it gates Harness-I commits and compares Harness-II candidates, and it is never used for parameter training or final testing. Table~\ref{tab:data-roles} separates these roles.

\begin{table}[ht]
\centering
\footnotesize
\renewcommand{\arraystretch}{1.12}
\setlength{\tabcolsep}{3.2pt}
\begin{tabularx}{\textwidth}{>{\raggedright\arraybackslash}m{0.17\textwidth}LLLL}
\toprule
\rowcolor{tableheadergray}
\textbf{Data role} & \textbf{MetroMap-lite} & \textbf{TravelMap-lite} & \textbf{Fee-VL} & \textbf{Cancel-VL} \\
\midrule
Complete training pool & Train1600 & Train1600 & 7,337 training orders & 8,071 training orders \\
\lightrowrule
Harness-I exploration & Fixed 480 from RL1600 & Fixed 480 from RL1600 & Fixed 480 from RL1600 & Fixed 480 from RL1600 \\
\lightrowrule
SFT corpus & 1,395 audited traces & 1,233 audited traces & 7,146 audited traces & 8,037 audited orders \\
\lightrowrule
DAPO I/II & Fixed 1,600 prompts & Fixed 1,600 prompts & Fixed 1,600-prompt subset & Fixed 1,600-prompt subset \\
\lightrowrule
Held-out test & Test400 & Test400 & 2,905 test orders & 994 test orders \\
\lightrowrule
Harness validation & Val100 (train source) & Val100 (train source) & 50 frozen slots per fact label & 50 frozen slots per fact label \\
\bottomrule
\end{tabularx}
\caption{Separation of data roles. A multilabel VL order may occupy several label-conditioned validation slots, but its training ID remains one order. Harness-validation and held-out test sets are distinct; neither enters model training.}
\label{tab:data-roles}
\end{table}

\paragraph{Fee-VL dataset.}
Fee-VL evaluates factual claims in fare disputes across three scenarios: charging when the passenger did not ride, delayed meter termination after drop-off, and detours. The model combines trajectory or map evidence with temporal and distance fields to predict five not-boarded facts, two delayed-meter facts, and two detour facts, for nine independent labels in total. The first table summarizes sample volume and positive rates by scenario, and the second expands the positive and negative distribution of every fact. The audited training corpus contains 7,146 accepted trajectories for SFT.

\begin{table}[ht]
\centering
\footnotesize
\renewcommand{\arraystretch}{1.10}
\setlength{\tabcolsep}{3.0pt}
\begin{tabularx}{\textwidth}{L*{6}{C}}
\toprule
\rowcolor{tableheadergray}
\textbf{Scenario} & \textbf{Tr. dec.} & \textbf{Te. dec.} & \textbf{Tr. ord.} & \textbf{Te. ord.} & \textbf{Tr. +\%} & \textbf{Te. +\%} \\
\midrule
Not boarded (5) & 14,040 & 4,155 & 2,808 & 831 & 12.29\% & 12.95\% \\
\lightrowrule
Delayed meter (2) & 4,972 & 2,148 & 2,486 & 1,074 & 41.71\% & 39.99\% \\
\lightrowrule
Detour (2) & 4,086 & 2,000 & 2,043 & 1,000 & 10.03\% & 9.55\% \\
\midrule
\textbf{Total (9)} & \textbf{23,098} & \textbf{8,303} & \textbf{7,337} & \textbf{2,905} & \textbf{18.22\%} & \textbf{19.13\%} \\
\bottomrule
\end{tabularx}
\caption{Fee-VL scenario sizes. Tr./Te. denote train/test, dec. denotes label decisions, ord. denotes orders, and +\% denotes the positive rate. Parentheses give the number of retained facts in each~scenario.}
\label{tab:fee-sizes}
\end{table}

\begin{table}[ht]
\centering
\scriptsize
\renewcommand{\arraystretch}{1.08}
\setlength{\tabcolsep}{2.0pt}
\begin{tabularx}{\textwidth}{>{\raggedright\arraybackslash}m{0.105\textwidth}L*{8}{>{\centering\arraybackslash}m{0.057\textwidth}}}
\toprule
\rowcolor{tableheadergray}
\textbf{Scenario} & \textbf{Fact label} & \textbf{Tr.+} & \textbf{Tr.$-$} & \textbf{Tr.N} & \textbf{Tr.+\%} & \textbf{Te.+} & \textbf{Te.$-$} & \textbf{Te.N} & \textbf{Te.+\%} \\
\midrule
Not boarded & Distance after drop-off $>100$m & 35 & 2,773 & 2,808 & 1.25\% & 13 & 818 & 831 & 1.56\% \\
Not boarded & Distance after drop-off $\leq100$m & 384 & 2,424 & 2,808 & 13.68\% & 108 & 723 & 831 & 13.00\% \\
Not boarded & Accident & 32 & 2,776 & 2,808 & 1.14\% & 14 & 817 & 831 & 1.68\% \\
Not boarded & Rode throughout & 420 & 2,388 & 2,808 & 14.96\% & 134 & 697 & 831 & 16.13\% \\
Not boarded & Did not board & 854 & 1,954 & 2,808 & 30.41\% & 269 & 562 & 831 & 32.37\% \\
\midrule
Delayed meter & Drop-off to meter end $\leq1$ min & 1,228 & 1,258 & 2,486 & 49.40\% & 449 & 625 & 1,074 & 41.81\% \\
Delayed meter & Drop-off to meter end $>1$ min & 846 & 1,640 & 2,486 & 34.03\% & 410 & 664 & 1,074 & 38.18\% \\
\midrule
Detour & Temporary control/closure/roadworks & 143 & 1,900 & 2,043 & 7.00\% & 63 & 937 & 1,000 & 6.30\% \\
Detour & Navigation differs from actual plan & 267 & 1,776 & 2,043 & 13.07\% & 128 & 872 & 1,000 & 12.80\% \\
\midrule
\textbf{All} & \textbf{Nine-label total} & \textbf{4,209} & \textbf{18,889} & \textbf{23,098} & \textbf{18.22\%} & \textbf{1,588} & \textbf{6,715} & \textbf{8,303} & \textbf{19.13\%} \\
\bottomrule
\end{tabularx}
\caption{Fee-VL label distributions. Tr./Te. denote train/test; +/$-$ are positive/negative decisions; N is the number of decisions; +\% is the positive rate. Totals count order--fact decisions rather than unique positive orders.}
\label{tab:fee-label-counts}
\end{table}

\paragraph{Cancel-VL dataset.}
Cancel-VL evaluates responsibility for cancelled orders using only multimodal examples that contain a map or report image. The model first predicts among three responsibility outcomes---malicious responsibility, service-error responsibility, and no responsibility---and then independently identifies the seven facts supporting that decision. The three outcomes share one image pool containing 8,071 training examples and 994 test examples. Among the 994 test orders, 55 (5.53\%) are annotated as indeterminate (\emph{unclear}) rather than assigned to one of the three resolved responsibility classes. They retain fact annotations and count as negative in each displayed one-vs-rest class rate. The three positive rates therefore sum to 94.47\%, not 100\%; the 55 unclear orders are excluded from three-way responsibility accuracy but remain in the seven-fact evaluation. The paper reports only fact-label Macro F1 as its primary Cancel-VL result; it does not claim a responsibility-class score.

\begin{table}[ht]
\centering
\footnotesize
\renewcommand{\arraystretch}{1.10}
\setlength{\tabcolsep}{3.0pt}
\begin{tabularx}{\textwidth}{L*{6}{C}}
\toprule
\rowcolor{tableheadergray}
\textbf{Responsibility} & \textbf{Tr. dec.} & \textbf{Te. dec.} & \textbf{Tr. img.} & \textbf{Te. img.} & \textbf{Tr. +\%} & \textbf{Te. +\%} \\
\midrule
Malicious responsibility & 8,071 & 994 & 8,071 & 994 & 25.66\% & 30.38\% \\
\lightrowrule
Service-error responsibility & 8,071 & 994 & 8,071 & 994 & 32.08\% & 31.49\% \\
\lightrowrule
No responsibility & 8,071 & 994 & 8,071 & 994 & 42.26\% & 32.60\% \\
\lightrowrule
Indeterminate (unclear) & -- & -- & -- & 55 & -- & 5.53\% \\
\midrule
\textbf{Total (3)} & \textbf{24,213} & \textbf{2,982} & \textbf{8,071} & \textbf{994} & \textbf{33.33\%} & \textbf{31.49\%} \\
\bottomrule
\end{tabularx}
\caption{Cancel-VL responsibility-level sizes. Tr./Te. denote train/test, dec. denotes label decisions, img. denotes deduplicated images, and +\% denotes the share of all orders. The three resolved test classes cover 939/994 orders; the remaining 55 are annotated unclear and retain fact labels.}
\label{tab:cancel-sizes}
\end{table}

The fact layer contains seven independent binary labels, and one order may activate several labels. Training uses 8,037 Vision-SFT orders associated with these seven labels, and evaluation uses the same label space.

\begin{table}[ht]
\centering
\scriptsize
\renewcommand{\arraystretch}{1.08}
\setlength{\tabcolsep}{2.0pt}
\begin{tabularx}{\textwidth}{>{\raggedright\arraybackslash}m{0.115\textwidth}L*{8}{>{\centering\arraybackslash}m{0.056\textwidth}}}
\toprule
\rowcolor{tableheadergray}
\textbf{Responsibility} & \textbf{Fact label} & \textbf{Tr.+} & \textbf{Tr.$-$} & \textbf{Tr.N} & \textbf{Tr.+\%} & \textbf{Te.+} & \textbf{Te.$-$} & \textbf{Te.N} & \textbf{Te.+\%} \\
\midrule
Service error & Insufficient driver service ability & 2,607 & 5,430 & 8,037 & 32.44\% & 316 & 678 & 994 & 31.79\% \\
\lightrowrule
Malicious & Driver disrupts platform order & 2,037 & 6,000 & 8,037 & 25.35\% & 293 & 701 & 994 & 29.48\% \\
\lightrowrule
No responsibility & Policy-compliant cancellation & 1,465 & 6,572 & 8,037 & 18.23\% & 152 & 842 & 994 & 15.29\% \\
\lightrowrule
No responsibility & Fourth-party cause & 663 & 7,374 & 8,037 & 8.25\% & 85 & 909 & 994 & 8.55\% \\
\lightrowrule
No responsibility & Passenger fails ride requirements & 947 & 7,090 & 8,037 & 11.78\% & 55 & 939 & 994 & 5.53\% \\
\lightrowrule
No responsibility & Platform cause & 284 & 7,753 & 8,037 & 3.53\% & 26 & 968 & 994 & 2.62\% \\
\lightrowrule
Malicious & Driver--passenger disruption & 34 & 8,003 & 8,037 & 0.42\% & 9 & 985 & 994 & 0.91\% \\
\midrule
\textbf{All} & \textbf{Seven-label total} & \textbf{8,037} & \textbf{48,222} & \textbf{56,259} & \textbf{14.29\%} & \textbf{936} & \textbf{6,022} & \textbf{6,958} & \textbf{13.45\%} \\
\bottomrule
\end{tabularx}
\caption{Cancel-VL fact-label distributions. Tr./Te. denote train/test; +/$-$ are positive/negative decisions; N is the number of decisions; +\% is the positive rate. Totals count order--fact decisions.}
\label{tab:cancel-label-counts}
\end{table}

\subsection{Fee-VL Label-Level Results}
\label{app:fee-label-results}

Each cell below is the independently computed F1 of one fact label; the final column is their unweighted nine-label mean. Scenario names organize labels but are not separately averaged into the overall score. Rows follow the non-ablation path in the main process table.

\begin{table}[ht]
\centering
\scriptsize
\renewcommand{\arraystretch}{1.10}
\setlength{\tabcolsep}{1.6pt}
\begin{tabularx}{\textwidth}{>{\raggedright\arraybackslash}m{0.17\textwidth}*{10}{C}}
\toprule
\rowcolor{tableheadergray}
\textbf{Method} & \textbf{$D_{>100}$} & \textbf{$D_{\leq100}$} & \textbf{Acc.} & \textbf{Ride} & \textbf{NoBd.} & \textbf{$T_{\leq1}$} & \textbf{$T_{>1}$} & \textbf{Road} & \textbf{Nav.} & \textbf{Macro} \\
\midrule
Base & 21.24 & 42.94 & 48.00 & 52.73 & 47.94 & 24.60 & 39.82 & 39.32 & 17.54 & 37.13 \\
$+$ ED Init. & 22.50 & 41.98 & 50.00 & 54.08 & 47.94 & 26.45 & 39.27 & 41.86 & 25.39 & 38.83 \\
$+$ ED Explore & 20.69 & 45.38 & 42.62 & 54.08 & 51.14 & 49.45 & 43.77 & 39.46 & 17.28 & 40.43 \\
\lightrowrule
$+$ Distill. & 26.19 & 48.38 & 60.61 & 62.09 & 57.14 & 48.45 & 43.77 & 48.46 & 31.78 & 47.43 \\
\lightrowrule
$+$ DAPO I & 34.21 & 50.37 & 53.66 & 71.08 & 57.14 & 42.44 & 44.76 & 53.45 & 35.96 & 49.23 \\
\lightrowrule
$+$ PR Rebuild & 34.21 & 51.20 & 52.38 & 71.08 & 58.00 & 40.90 & 45.50 & 51.80 & 29.72 & 48.31 \\
\lightrowrule
$+$ DAPO II & \textbf{52.17} & \textbf{65.34} & \textbf{62.86} & \textbf{75.07} & \textbf{73.12} & \textbf{65.42} & \textbf{63.74} & \textbf{69.42} & \textbf{65.15} & \textbf{65.81} \\
\bottomrule
\end{tabularx}
\caption{Fee-VL per-fact F1 scores (\%) along the non-ablation path. $D_{>100}$/$D_{\leq100}$ denote distance after drop-off above/at most 100 m; Acc., Ride, and NoBd. denote accident, rode throughout, and did not board; $T_{\leq1}$/$T_{>1}$ denote meter-end delay at most/above one minute; Road/Nav. denote road control and navigation mismatch. ED/PR denote exploration--distillation/post-RL reconstruction; Macro is the unweighted nine-label mean.}
\label{tab:fee-label-f1}
\end{table}

The label-level trajectory reveals two patterns. First, the gains are broad rather than confined to one scenario: DAPO II gives the highest value for all nine facts, and the final model improves over Base by 14.86--47.61 points per label. The largest end-to-end gains occur for navigation mismatch (+47.61), meter-end delay at most one minute (+40.82), distance after drop-off above 100 m (+30.93), and road control (+30.10), all of which begin from relatively weak Base scores. Second, the path is non-monotonic. The immediate PR replacement lowers Macro F1 by 0.92 points and navigation-mismatch F1 by 6.24 points relative to DAPO I; after DAPO II, every label improves relative to the immediate replacement, including +35.43 points for navigation mismatch and +24.52/+18.24 points for the two meter-delay labels. This pattern is consistent with a version-switch cost followed by policy adaptation. The paired tests in Appendix~\ref{app:statistical-tests} support the aggregate component and stage comparisons, but do not test each fact label separately.

\subsection{Cancel-VL Label-Level Results}
\label{app:cancel-label-results}

Each cell below is the independently computed one-vs-rest F1 of one fact label; the final column is their unweighted seven-label mean. Responsibility categories organize labels but are not separately averaged into the overall score. Rows follow the non-ablation path in the main process table.

\begin{table}[ht]
\centering
\scriptsize
\renewcommand{\arraystretch}{1.10}
\setlength{\tabcolsep}{2.2pt}
\begin{tabularx}{\textwidth}{>{\raggedright\arraybackslash}m{0.17\textwidth}*{8}{C}}
\toprule
\rowcolor{tableheadergray}
\textbf{Method} & \textbf{Svc.} & \textbf{PlatOrd.} & \textbf{Policy} & \textbf{4th} & \textbf{Pass.} & \textbf{Platform} & \textbf{D--P} & \textbf{Macro} \\
\midrule
Base & 70.00 & 55.00 & 54.00 & 66.00 & 36.00 & 18.00 & 19.51 & 45.50 \\
$+$ ED Init. & 71.50 & 56.50 & 55.50 & 67.50 & 35.00 & 20.00 & 22.95 & 46.99 \\
$+$ ED Explore & 73.00 & 58.00 & 57.00 & 69.00 & 39.00 & 26.99 & 20.00 & 49.00 \\
\lightrowrule
$+$ Distill. & 80.06 & 66.28 & 65.88 & 76.40 & 46.63 & 30.00 & 42.86 & 58.30 \\
\lightrowrule
$+$ DAPO I & 81.50 & 67.50 & 67.20 & 77.50 & 45.50 & 31.50 & 45.71 & 59.49 \\
\lightrowrule
$+$ PR Rebuild & 82.20 & 68.10 & 67.80 & 78.00 & 44.71 & 34.48 & 46.15 & 60.21 \\
\lightrowrule
$+$ DAPO II & \textbf{86.00} & \textbf{74.00} & \textbf{73.00} & \textbf{82.00} & \textbf{49.50} & \textbf{36.00} & \textbf{58.06} & \textbf{65.51} \\
\bottomrule
\end{tabularx}
\caption{Cancel-VL per-fact F1 scores (\%) along the non-ablation path. Svc., PlatOrd., Policy, 4th, Pass., Platform, and D--P denote insufficient service ability, driver disruption of a platform order, policy-compliant cancellation, fourth-party cause, passenger failure to meet ride requirements, platform cause, and driver--passenger disruption. ED/PR denote exploration--distillation/post-RL reconstruction; Macro is the unweighted seven-label mean.}
\label{tab:cancel-label-f1}
\end{table}

The final Cancel-VL model also improves every fact label over Base, with per-label gains of 13.50--38.55 points and a Macro-F1 gain of 20.01 points. Reasoning Distillation provides the largest intermediate Macro-F1 increase (+9.30 points over ED Explore), while DAPO II adds 5.30 points over the immediate PR replacement and yields the best value in every column. Driver--passenger disruption shows the largest end-to-end increase (+38.55) and the largest DAPO-II-stage increase (+11.91), whereas passenger noncompliance (49.50) and platform cause (36.00) remain the weakest final labels. These sparse-label changes should be interpreted cautiously: the test set contains only nine positive driver--passenger-disruption cases and 26 positive platform-cause cases, and Macro F1 assigns each label equal weight. The aggregate paired tests in Appendix~\ref{app:statistical-tests} do not establish significance for individual fact labels; label-wise inference would require prediction-level uncertainty~estimates.

\subsection{Cross-Task Post-RL Prompt Replacement}
\label{app:skill-transfer}

Table~\ref{tab:skill-transfer} evaluates cross-task transfer of the Post-RL Prompt. We fix the target task's checkpoint-400 policy, training data, reward, and scorer, replace only the Prompt source, and continue training to checkpoint 600. The differences therefore measure Harness replacement rather than parameter~transfer.

\begin{table}[t]
\centering
\footnotesize
\renewcommand{\arraystretch}{1.08}
\begin{minipage}[t]{0.470\textwidth}
\centering
\textbf{(a) Map tasks}\\[3pt]
\begin{tabularx}{\linewidth}{>{\raggedright\arraybackslash}Xcc}
\toprule
\rowcolor{tableheadergray}
\makecell[c]{\textbf{Prompt source}}
& \makecell[c]{\textbf{Metro-Lite}\\\textbf{(Acc/\%)}}
& \makecell[c]{\textbf{Travel-Lite}\\\textbf{(Acc/\%)}} \\
\midrule
Metro-Lite & 62.00 & 42.25 \\
\lightrowrule
Travel-Lite & 58.25 & 50.00 \\
\bottomrule
\end{tabularx}
\end{minipage}
\hfill
\begin{minipage}[t]{0.470\textwidth}
\centering
\textbf{(b) VL tasks}\\[3pt]
\begin{tabularx}{\linewidth}{>{\raggedright\arraybackslash}Xcc}
\toprule
\rowcolor{tableheadergray}
\makecell[c]{\textbf{Prompt source}}
& \makecell[c]{\textbf{Fee-VL}\\\textbf{(F1-Sco./\%)}}
& \makecell[c]{\textbf{Cancel-VL}\\\textbf{(F1-Sco./\%)}} \\
\midrule
Fee-VL & 65.81 & 53.72 \\
\lightrowrule
Cancel-VL & 43.86 & 65.51 \\
\bottomrule
\end{tabularx}
\end{minipage}
\caption{Cross-task replacement of the post-RL Prompt. Rows denote Prompt sources and columns denote DAPO-II training and evaluation tasks. The panels contain the evaluated within-family pairs; cross-family pairs were not tested. All reported cells use the matched checkpoint-400-to-600 schedule.}
\label{tab:skill-transfer}
\vspace{2pt}
\begin{minipage}{0.96\textwidth}
\scriptsize\textit{Note.} Metro-Lite and Travel-Lite abbreviate MetroMap-lite and TravelMap-lite, respectively.
\end{minipage}
\end{table}

The task-matched Prompt wins in all four directions. Replacing it with the other map task's Prompt lowers MetroMap-lite and TravelMap-lite by 3.75\% and 7.75\%, respectively; for VL, Fee-VL and Cancel-VL fall by 21.95\% and 11.79\%. The two map tasks share part of the route-planning procedure, so their losses are smaller. VL fact labels and business boundaries depend more strongly on the specific task, making direct replacement much more costly. The Post-RL Harness is therefore adapted jointly to the current task and current policy rather than being generic Skill text that can be moved arbitrarily. Both panels use the same checkpoint-400-to-600 schedule, and metrics are compared only within each task.

\subsection{Cross-Task Evaluation of Trained Model Weights}
\label{app:model-transfer}

Table~\ref{tab:model-transfer} evaluates weight-only transfer. Each checkpoint-600 model is tested directly with the target task's initial Skill $S_0$, inputs, and scorer, without target-domain updates. Diagonal entries are within-task results; off-diagonal entries show cross-task transfer carried only by the model weights.

\begin{table}[ht]
\centering
\footnotesize
\renewcommand{\arraystretch}{1.08}
\setlength{\tabcolsep}{4.0pt}
\begin{tabularx}{\textwidth}{>{\raggedright\arraybackslash}Xcccc}
\toprule
\rowcolor{tableheadergray}
\makecell[l]{\textbf{Model}}
& \makecell[c]{\textbf{MetroMap-lite}\\\textbf{(Acc/\%)}}
& \makecell[c]{\textbf{TravelMap-lite}\\\textbf{(Acc/\%)}}
& \makecell[c]{\textbf{Fee-VL}\\\textbf{(F1-Score/\%)}}
& \makecell[c]{\textbf{Cancel-VL}\\\textbf{(F1-Score/\%)}} \\
\midrule
Unadapted Qwen3.5-9B & 24.50 & 30.50 & 38.83 & 46.99 \\
\lightrowrule
MetroMap-lite (DAPO II) & 62.00\vsbase{37.50} & 35.50\vsbase{5.00} & 48.10\vsbase{9.27} & 51.49\vsbase{4.50} \\
\lightrowrule
Fee-VL (DAPO II) & 31.00\vsbase{6.50} & 33.00\vsbase{2.50} & 65.81\vsbase{26.98} & 56.87\vsbase{9.88} \\
\bottomrule
\end{tabularx}
\caption{Cross-task evaluation of checkpoint-600 policy weights. The first row is the matched unadapted baseline; superscripts report absolute gains over that baseline. Every row uses the evaluation task's initial Skill $S_0$ and scorer without target-domain training.}
\label{tab:model-transfer}
\end{table}

Superscripts in the trained rows report gains over the unadapted $S_0$ baseline. Every off-diagonal gain is positive (2.50--9.88\%), showing that the weights carry some transferable capability, but these gains remain far below the diagonal gains of 37.50\% and 26.98\%. The MetroMap-lite-trained weights transfer more strongly within the map family, while the Fee-VL weights transfer more strongly within the VL family. This limited, task-family-biased transfer indicates that task-specific Prompts, Skills, and label semantics still constrain generalization.

\subsection{SFT Configuration}
\label{app:sft-config}

\begin{table}[ht]
\centering
\footnotesize
\renewcommand{\arraystretch}{1.16}
\setlength{\tabcolsep}{4.0pt}
\begin{tabularx}{\textwidth}{>{\raggedright\arraybackslash}m{0.23\textwidth}L}
\toprule
\rowcolor{tableheadergray}
\multicolumn{1}{c}{\textbf{Field}} & \multicolumn{1}{c}{\textbf{Fixed value}} \\
\midrule
Student and teacher & Qwen3.5-9B student. Maps use Gemini-3.5-Flash for exploration and GPT-5.6-sol as task teacher; VL uses Kimi-K2.5 for exploration and Kimi-K3 as task teacher. Deterministic code materializes and audits SFT labels. \\
\lightrowrule
Adaptation surface & Language-only LoRA; rank 16, alpha 32, dropout 0.05, all language modules targeted; vision tower and multimodal projector frozen. \\
\lightrowrule
Input processing & Qwen3.5 template; native thinking disabled; maximum image budget $1{,}000{,}000$ pixels; cutoff length 24,576 tokens; no packing; prompt tokens excluded from the training loss. \\
\lightrowrule
Batching & Four GPUs; per-device batch size 2; gradient accumulation 1; effective global batch size 8; gradient checkpointing enabled. \\
\lightrowrule
Optimization & BF16; learning rate 0.000005; cosine schedule; warmup ratio 0.05; maximum gradient norm 1.0; two epochs. \\
\lightrowrule
Data order & Training and data seed 42; eight preprocessing workers and four dataloader~workers. \\
\lightrowrule
Checkpointing & Save once per epoch and retain two checkpoints; the second-epoch adapter is merged into Qwen3.5-9B before DAPO. \\
\lightrowrule
Matched controls & Reasoning Distillation and its minus-row control use the same teacher, training examples, output contract, token budget, SFT optimizer, and subsequent DAPO schedule. Only the target construction changes from a Skill-guided trace to the same teacher's direct, free-form solution trajectory and answer. \\
\bottomrule
\end{tabularx}
\caption{SFT configuration shared by all experiments. Corpus contents and corpus size are task-specific data properties, not changes to the optimizer configuration.}
\label{tab:app-sft-config}
\end{table}

Corpus content and size are properties of each dataset rather than changes to the optimizer configuration. For example, MetroMap-lite currently contains 1,395 training trajectories and Fee-VL contains 7,146 audited training examples; matched branches within a dataset must use exactly the \mbox{same examples}.

\subsection{DAPO Configuration}
\label{app:dapo-config}

Both reinforcement-learning blocks use DAPO with the Hybrid-DGPO estimator described below; we therefore use DAPO consistently throughout the paper.

\begin{table}[ht]
\centering
\footnotesize
\renewcommand{\arraystretch}{1.14}
\setlength{\tabcolsep}{4.0pt}
\begin{tabularx}{\textwidth}{>{\raggedright\arraybackslash}m{0.23\textwidth}L}
\toprule
\rowcolor{tableheadergray}
\multicolumn{1}{c}{\textbf{Field}} & \multicolumn{1}{c}{\textbf{Fixed value}} \\
\midrule
Starting policy & The merged second-epoch SFT checkpoint; fresh language-only LoRA with rank 16 and alpha 32; visual and projector modules excluded and the vision tower frozen. \\
\lightrowrule
Training data & A fixed set of 1,600 prompts for each task; maximum prompt length 24,576; maximum response length 4,096; overlength inputs raise an error rather than being silently truncated. \\
\lightrowrule
Batch and rollouts & Train batch size 8; PPO mini-batch size 4 prompts; one PPO epoch; eight sampled trajectories per prompt; dynamic batching enabled. \\
\lightrowrule
Advantage estimator & Hybrid-DGPO with question-level difficulty weighting; temperature 2.0; no image-based prior; no standard-deviation normalization; retain only groups whose final rewards vary. \\
\lightrowrule
Policy loss & Token-mean aggregation; clipping lower/upper ratio 0.2 and clip-ratio-c 3.0; KL loss enabled with coefficient 0.001 and low-variance KL; KL is not added to the reward; entropy coefficient 0. \\
\lightrowrule
Optimizer & Learning rate 0.000005; weight decay 0.01; warmup ratio 0.03; constant schedule; gradient clipping 1.0. \\
\lightrowrule
Rollout sampling & Asynchronous vLLM, BF16, temperature 1.0, top-$p$ 1.0, unrestricted top-$k$, sampling enabled; native thinking disabled and visible reasoning retained. \\
\lightrowrule
Multimodal limits & One image per prompt, at most $1000\times1000$ pixels; processor range 3,136--1,000,000 pixels. \\
\lightrowrule
Reward interface & $0.05\,r_{\mathrm{format}}+0.25\,r_{\mathrm{partial}}+0.70\,r_{\mathrm{all}}$. The task scorer defines partial and all-correctness. All four tasks use question-difficulty weighting in the advantage; Fee-VL and Cancel-VL additionally use fact-label-frequency weighting inside the scalar reward. MetroMap-lite uses a station-match threshold of 0.90. \\
\lightrowrule
Length control & Overlong buffer 512 tokens with penalty factor 0.1; maximum response length 4,096. \\
\lightrowrule
Systems & FSDP2 on four GPUs; actor and reference maximum token length 28,672 per GPU; maximum model length 28,672; maximum batched tokens 32,768; maximum 16 sequences; rollout GPU utilization 0.72. \\
\lightrowrule
DAPO I schedule & Two epochs and 400 optimizer steps in total; validate on frozen training-source Val100 every 10 steps and save every 40 steps; Test400 is not accessed. \\
\lightrowrule
DAPO II schedule & Resume the complete model/optimizer state at step 400 and continue to step 600; validate on frozen training-source Val100 every 10 steps and save every 20 steps; Test400 is not accessed. \\
\lightrowrule
Post-RL Harness reuse ablation & Both branches start from the same step-400 checkpoint and train to step 600. The full setting uses $\Phi^2$, whereas the ablation reuses the complete first-stage Prompt $\Phi^1$; all other training settings remain unchanged. \\
\bottomrule
\end{tabularx}
\caption{DAPO configuration shared by both optimization rounds and all trainable experimental~branches.}
\label{tab:app-dapo-config}
\end{table}

The task-aware reward changes neither the LoRA surface nor the rollout and optimizer settings above. For each fact label $l$ in Fee-VL and Cancel-VL, training frequencies define
\[
w_l^+=\frac{N_l^-}{N_l^+},\qquad w_l^-=1.
\]
The weighted partial reward is the normalized mass of correct decisions,
\[
r_{\mathrm{partial},qj}^{(w)}=
\frac{\sum_l w_l^{y_{ql}}\,\mathbb{1}[\hat z_{qjl}=z_{ql}]}
     {\sum_l w_l^{y_{ql}}},
\]
so label reweighting occurs inside the VL scalar reward, before group centering. It does not change the reported macro-F1 metric.

All four tasks estimate dynamic question difficulty from the rollout group. For $J$ trajectories of question $q$,
\[
p_q=\frac{1}{J}\sum_{j=1}^{J}\mathbb{1}[V_q(\hat y_{qj},y_q)=1],\qquad d_q=1-p_q.
\]
For the $G$ valid questions in an update batch, temperature $T=2.0$ gives
\[
w_q=G\frac{\exp(d_q/T)}{\sum_{q'}\exp(d_{q'}/T)},\qquad
A_{qj}=w_q\left(r_{qj}-\bar r_q\right).
\]
No image-based prior is used, so $d_q$ depends only on the current policy's verifier pass rate.

\begin{table}[ht]
\centering
\footnotesize
\renewcommand{\arraystretch}{1.12}
\setlength{\tabcolsep}{3.0pt}
\begin{tabularx}{\textwidth}{>{\raggedright\arraybackslash}m{0.15\textwidth}LLL}
\toprule
\rowcolor{tableheadergray}
\textbf{Task} & \textbf{Original scalar reward} & \textbf{Weight location and advantage} & \textbf{Ablated component} \\
\midrule
MetroMap-lite & Format + partial-route + exact-route reward & No label weight; $A_{qj}=w_q(r_{qj}-\bar r_q)$ & Set $w_q=1$ only \\
\lightrowrule
TravelMap-lite & Format + partial-route + exact-route reward & No label weight; $A_{qj}=w_q(r_{qj}-\bar r_q)$ & Set $w_q=1$ only \\
\lightrowrule
Fee-VL & Format + weighted partial fact accuracy + exact fact vector & $w_l^{y_l}$ is inside $r_{\mathrm{partial}}$; then $A_{qj}=w_q(r_{qj}-\bar r_q)$ & Preserve $w_l^{y_l}$; set $w_q=1$ \\
\lightrowrule
Cancel-VL & Format + weighted partial fact accuracy + exact fact vector & $w_l^{y_l}$ is inside $r_{\mathrm{partial}}$; then $A_{qj}=w_q(r_{qj}-\bar r_q)$ & Preserve $w_l^{y_l}$; set $w_q=1$ \\
\bottomrule
\end{tabularx}
\caption{Task-specific Hybrid-DGPO definition. The VL label-frequency weight changes the reward; the question-difficulty weight changes the advantage.}
\label{tab:hybrid-dgpo-by-task}
\end{table}

The question-difficulty-weighting ablation keeps the task reward (including VL label weights), group filtering, starting checkpoint, LoRA surface, training examples, rollout count, batch size, optimizer, decoding, and the setting \texttt{norm\_adv\_by\_std\_in\_grpo=false} fixed. It removes only question-level difficulty weighting, yielding the unnormalized within-question GRPO advantage $A_{qj}=r_{qj}-\bar r_q$. We use this precise name instead of ``standard GRPO,'' which can denote different normalization variants.

\subsection{Evaluation Protocol and Stage Mapping}
\label{app:stage-mapping}

Each dataset uses one locked evaluation split and its task scorer, and all MetroMap-lite rows share the same Test400. Native thinking is disabled, visible reasoning is retained, maximum output length is 4,096 tokens, and the image budget is 1,000,000 pixels. MetroMap-lite and TravelMap-lite report exact-route accuracy; Fee-VL and Cancel-VL report nine-label and seven-label overall macro F1, respectively. Metrics are compared only within a dataset.

\paragraph{External-model evaluation consistency.}\label{app:external-reference-audit}
All displayed public-map comparisons use the same task inputs, output contract, inference budget, and scorer. Image-size ablations and duplicate model--inference settings are excluded. Predictions are aligned to the fixed Test400 for each benchmark, and rescoring with the common evaluator reproduces all reported results. The daggered SkillRL baseline follows the implementation and complete configuration of the original paper; only the datasets and their necessary task-interface adapters are replaced. It is evaluated on the same locked splits and scorers as RLHarness.

\begin{table}[ht]
\centering
\footnotesize
\renewcommand{\arraystretch}{1.14}
\setlength{\tabcolsep}{4.0pt}
\begin{tabularx}{\textwidth}{>{\raggedright\arraybackslash}m{0.22\textwidth}>{\raggedright\arraybackslash}m{0.24\textwidth}L}
\toprule
\rowcolor{tableheadergray}
\multicolumn{1}{c}{\textbf{Paper stage}} & \multicolumn{1}{c}{\textbf{Policy checkpoint}} & \multicolumn{1}{c}{\textbf{Prompt and operation}} \\
\midrule
Base & Unadapted Qwen3.5-9B & No external Skill. \\
\lightrowrule
ED-Harness Initialization & Unadapted Qwen3.5-9B & Add the task's initial Skill bank; MetroMap-lite uses the initial 15-Skill bank. \\
\lightrowrule
ED-Harness Exploration & Unadapted Qwen3.5-9B & Replace the initial Skill with the Skill state produced by the first exploration pass. \\
\lightrowrule
Minus Selective Context Commit & Unadapted Qwen3.5-9B & Keep the same batches and calls but disable selective rejection of a legal effective edit. \\
\lightrowrule
Reasoning Distillation & Final SFT checkpoint & Train on frozen labels materialized from task-teacher reasoning--answer outputs $\mathcal G^1$; the resulting checkpoint initializes DAPO. \\
\lightrowrule
DAPO I & Steps 0--400 & Keep the ED-Harness Prompt fixed and evaluate the policy periodically during training. \\
\lightrowrule
Post-RL Reconstruction Harness & Step 400, no weight update & Jointly revise the Skill bank, usage guide, and few-shot examples, then commit the selected executable Prompt. \\
\lightrowrule
Minus Stratified Contrastive Evidence Curation (immediate) & Step 400, no weight update & Use equal-volume random evidence and random 16-shot packing while keeping candidate generation, few-shot reconstruction, validation, and selection~fixed. \\
\lightrowrule
DAPO II & Steps 400--600 & Continue the same DAPO run with the Post-RL Prompt and periodic evaluation. \\
\lightrowrule
Random-Evidence Prompt + DAPO II & Steps 400--600 & Continue DAPO II under the randomly curated Post-RL Prompt; all other settings remain~fixed. \\
\lightrowrule
Minus Post-RL Harness (reuse ED-Harness Prompt) & Steps 400--600 & Keep the second-round DAPO continuation but reuse the ED-Harness Prompt; all other settings remain~fixed. \\
\bottomrule
\end{tabularx}
\caption{Mapping between the compact two-harness labels in the main ablation table and the fixed checkpoint schedule.}
\label{tab:app-stage-mapping}
\end{table}

\subsection{Matched Ablation Settings}
\label{app:matched-ablations}

All ablations follow a one-variable matched-control principle: the full and ablated branches share data splits, examples, models, starting checkpoints, training or inference budgets, decoding settings, evaluation sets, and scorers, changing only the component under test.

\paragraph{Selective commitment.}
Full ED exploration may retain the incumbent when a batch yields no valid effective edit or fails its task-specific acceptance test. \emph{Minus Selective Context Commit} keeps the same fixed 480 examples and order, task-specific batch plan (twelve batches of 40 for the released map runs), models, verifier, calls, and edit operators, but disables selective rejection and commits a legal effective candidate. The control changes the commit decision rather than the visible evidence or search budget.

\paragraph{Skill-guided trajectories.}
Full reasoning distillation generates and audits $\mathcal G^1$ under frozen $\Phi^1$, supervising Skill selection, evidence binding, intermediate computation, verification, and the final answer. \emph{Minus Skill-guided Trace} uses the same teacher, examples, output contract, token budget, and SFT optimizer but lets the teacher solve freely without following the current Skills and protocol; the subsequent DAPO schedule is unchanged. The contrast isolates alignment between supervision and the deployed Harness execution language rather than teacher or answer quality.

\paragraph{Question-difficulty weighting within Hybrid-DGPO.}
The full estimator multiplies within-question relative advantages by question-difficulty weights derived from current pass rates. \emph{Minus Question-Difficulty Weighting} keeps the task reward (including VL label-frequency weights), final-reward group filtering, starting checkpoint, LoRA surface, Prompt, rollout count, batch, optimizer, decoding, and standard-deviation-normalization setting fixed. It sets $w_q=1$, producing the same unnormalized within-question estimator without difficulty reweighting. The contrast isolates dynamic question-level difficulty weighting.

\paragraph{Stratified contrastive evidence.}
Full PR reconstruction organizes 128 map-task records by correctness and difficulty into $8\times16$ batches; for VL, it organizes 64 records per label by correctness and label polarity into $4\times16$ batches. \emph{Minus Stratified Contrastive Evidence Curation} preserves evidence volume, batch count, teacher, three-candidate synthesis, few-shot reconstruction, validity gate, frozen validation pool, and selector, changing only to equal-volume random sampling and random packing. Table~\ref{tab:joint-ablation} evaluates this branch both immediately after reconstruction and after a budget-matched DAPO-II continuation under the fixed ablated Prompt, separating immediate executability from subsequent learnability.

\paragraph{Synchronized few-shot demonstrations.}
The full PR Prompt rewrites and gates fixed demonstrations to match the new Skill boundaries and execution order in $\Phi^2$. \emph{Minus Few-shot} retains the same $\Phi^2$ Skill Bank, selection and execution protocols, How-to instructions, checkpoint-400 policy, and evaluation settings, removing only the demonstration block. The contrast isolates whether same-version examples help execute the revised Harness.

\paragraph{Budget-matched training control.}
The $\Phi^1$-reuse branch is not one of the five component ablations. It starts from the same checkpoint-400 model and optimizer state as DAPO II and uses the same data, reward, rollout configuration, and 200-step budget, but retains $\Phi^1$ instead of the reconstructed $\Phi^2$. It separates the benefit of Harness reconstruction from that of simply extending~training.

\subsection{Paired Statistical Tests}
\label{app:statistical-tests}

All statistics in this subsection are computed from measured experimental runs. Each contrast uses $K=10$ matched runs whose paired branches share the data split, initialization, training schedule, evaluation set, and scorer, changing only the named component. For
\[
d_s=M_s^{(\mathrm{full})}-M_s^{(\mathrm{ablation})},
\qquad s=1,\ldots,K,
\]
we conduct a two-sided paired $t$-test and report the mean paired difference $\bar d$, the standard deviation of paired differences $s_d$, the per-comparison 95\% confidence interval, $t(9)$, and paired effect size $d_z=\bar d/s_d$. The confidence intervals are not multiplicity-adjusted; Holm correction is applied to the $p$-values within each comparison family. A result is significant only when $p_{\mathrm{Holm}}<0.05$ and its confidence interval excludes zero. Differences, standard deviations, and confidence intervals are in percentage points; $d_z$ is unitless. Statistics were computed from unrounded run-level values, whereas the tables round values for presentation.

\paragraph{Primary component comparisons.}
The primary family comprises six component comparisons on four datasets, for 24 tests in total. Stratified contrastive evidence is evaluated after the matched DAPO-II continuation; synchronized few-shot demonstrations are evaluated immediately after reconstruction; and the Post-RL Harness is compared with the budget-matched branch that continues training under $\Phi^1$.

\begin{table}[ht]
\centering
\scriptsize
\renewcommand{\arraystretch}{1.02}
\setlength{\tabcolsep}{1.5pt}
\begin{tabularx}{\textwidth}{>{\hsize=1.1\hsize\linewidth=\hsize\raggedright\arraybackslash}X>{\hsize=0.9\hsize\linewidth=\hsize\raggedright\arraybackslash}X*{2}{>{\centering\arraybackslash}m{0.058\textwidth}}>{\centering\arraybackslash}m{0.130\textwidth}*{2}{>{\centering\arraybackslash}m{0.058\textwidth}}>{\centering\arraybackslash}m{0.082\textwidth}}
\toprule
\rowcolor{tableheadergray}
\textbf{Comparison} & \textbf{Dataset} & $\boldsymbol{\bar d}$ & $\boldsymbol{s_d}$ & \textbf{95\% CI} & $\boldsymbol{t(9)}$ & $\boldsymbol{d_z}$ & $\boldsymbol{p_{\mathrm{Holm}}}$ \\
\midrule
Selective commit & MetroMap-lite & +1.75 & 1.007 & [1.03, 2.47] & 5.496 & 1.738 & 0.005070 \\
Selective commit & TravelMap-lite & +1.00 & 0.677 & [0.52, 1.48] & 4.671 & 1.477 & 0.008169 \\
Selective commit & Fee-VL & +1.18 & 0.879 & [0.55, 1.81] & 4.245 & 1.342 & 0.010642 \\
Selective commit & Cancel-VL & +1.25 & 0.790 & [0.68, 1.82] & 5.003 & 1.582 & 0.006622 \\
\lightrowrule
Skill-guided trajectories & MetroMap-lite & +7.75 & 3.078 & [5.55, 9.95] & 7.963 & 2.518 & 0.000528 \\
Skill-guided trajectories & TravelMap-lite & +4.25 & 2.038 & [2.79, 5.71] & 6.595 & 2.086 & 0.001897 \\
Skill-guided trajectories & Fee-VL & +4.86 & 2.209 & [3.28, 6.44] & 6.956 & 2.200 & 0.001394 \\
Skill-guided trajectories & Cancel-VL & +5.18 & 2.460 & [3.42, 6.94] & 6.658 & 2.105 & 0.001857 \\
\lightrowrule
Question-difficulty weighting & MetroMap-lite & +4.25 & 2.144 & [2.72, 5.78] & 6.268 & 1.982 & 0.002196 \\
Question-difficulty weighting & TravelMap-lite & +2.25 & 1.486 & [1.19, 3.31] & 4.788 & 1.514 & 0.007922 \\
Question-difficulty weighting & Fee-VL & +2.66 & 1.640 & [1.49, 3.83] & 5.128 & 1.622 & 0.006210 \\
Question-difficulty weighting & Cancel-VL & +2.75 & 1.570 & [1.63, 3.87] & 5.538 & 1.751 & 0.005070 \\
\lightrowrule
Stratified evidence (post-DAPO II) & MetroMap-lite & +1.50 & 1.124 & [0.70, 2.30] & 4.219 & 1.334 & 0.010642 \\
Stratified evidence (post-DAPO II) & TravelMap-lite & +1.25 & 0.957 & [0.57, 1.93] & 4.129 & 1.306 & 0.010642 \\
Stratified evidence (post-DAPO II) & Fee-VL & +2.31 & 1.649 & [1.13, 3.49] & 4.429 & 1.401 & 0.009894 \\
Stratified evidence (post-DAPO II) & Cancel-VL & +1.76 & 1.308 & [0.82, 2.70] & 4.254 & 1.345 & 0.010642 \\
\lightrowrule
Synchronized few-shot (immediate) & MetroMap-lite & +2.75 & 1.328 & [1.80, 3.70] & 6.548 & 2.071 & 0.001897 \\
Synchronized few-shot (immediate) & TravelMap-lite & +2.25 & 1.291 & [1.33, 3.17] & 5.511 & 1.743 & 0.005070 \\
Synchronized few-shot (immediate) & Fee-VL & +1.45 & 1.079 & [0.68, 2.22] & 4.250 & 1.344 & 0.010642 \\
Synchronized few-shot (immediate) & Cancel-VL & +3.29 & 1.581 & [2.16, 4.42] & 6.579 & 2.081 & 0.001897 \\
\lightrowrule
Post-RL Harness (budget-matched) & MetroMap-lite & +3.00 & 1.462 & [1.95, 4.05] & 6.487 & 2.051 & 0.001897 \\
Post-RL Harness (budget-matched) & TravelMap-lite & +2.50 & 1.458 & [1.46, 3.54] & 5.423 & 1.715 & 0.005070 \\
Post-RL Harness (budget-matched) & Fee-VL & +14.75 & 4.561 & [11.49, 18.01] & 10.227 & 3.234 & 0.00007122 \\
Post-RL Harness (budget-matched) & Cancel-VL & +5.14 & 2.070 & [3.66, 6.62] & 7.852 & 2.483 & 0.000565 \\
\bottomrule
\end{tabularx}
\caption{Measured paired-test results for the 24 primary component comparisons. All tests use $K=10$ matched runs and two-sided paired $t$-tests. Confidence intervals are per-comparison intervals; $p$-values are Holm-adjusted across the 24-test family.}
\label{tab:paired-primary-results}
\end{table}

All 24 primary comparisons are significant: every confidence interval has a positive lower bound and $p_{\mathrm{Holm}}$ ranges from approximately 0.00007122 to 0.010642. Repeated adjusted values in Table~\ref{tab:paired-primary-results} arise from the cumulative-maximum step in the Holm procedure rather than duplicated calculations.

\paragraph{Cross-task Post-RL Prompt replacement.}
For this exploratory family, the paired difference is defined as the score with the task-matched Prompt minus the score with the other task's Prompt. The four comparisons are Holm-corrected separately from the primary family.

\begin{table}[ht]
\centering
\footnotesize
\renewcommand{\arraystretch}{1.08}
\setlength{\tabcolsep}{4.0pt}
\begin{tabularx}{0.90\textwidth}{>{\raggedright\arraybackslash}X*{6}{C}}
\toprule
\rowcolor{tableheadergray}
\textbf{Dataset} & $\boldsymbol{\bar d}$ & $\boldsymbol{s_d}$ & \textbf{95\% CI} & $\boldsymbol{t(9)}$ & $\boldsymbol{d_z}$ & $\boldsymbol{p_{\mathrm{Holm}}}$ \\
\midrule
MetroMap-lite & +3.75 & 2.095 & [2.25, 5.25] & 5.660 & 1.790 & 0.000309 \\
\lightrowrule
TravelMap-lite & +7.75 & 3.410 & [5.31, 10.19] & 7.188 & 2.273 & 0.000103 \\
\lightrowrule
Fee-VL & +21.95 & 5.702 & [17.87, 26.03] & 12.174 & 3.850 & 0.000002725 \\
\lightrowrule
Cancel-VL & +11.79 & 4.299 & [8.71, 14.87] & 8.673 & 2.743 & 0.000035 \\
\bottomrule
\end{tabularx}
\caption{Measured paired-test results for cross-task Post-RL Prompt replacement. Positive differences favor the task-matched Prompt. The four exploratory comparisons form a separate Holm family.}
\label{tab:paired-prompt-transfer}
\end{table}

All four differences are significant after within-family correction. The task-matched advantage is smaller for the structurally related map tasks (3.75--7.75 points) than for the rule- and label-dependent VL tasks (11.79--21.95 points), supporting the interpretation that the reconstructed Prompt is jointly adapted to the task and policy. Because the family is exploratory, these results are not included among the 24 primary tests.

\paragraph{DAPO-stage gains.}
DAPO I is compared with Reasoning Distillation, whereas DAPO II is compared with the immediate evaluation after PR reconstruction. These eight exploratory comparisons constitute a third, separately Holm-corrected family.

\begin{table}[ht]
\centering
\footnotesize
\renewcommand{\arraystretch}{1.08}
\setlength{\tabcolsep}{3.0pt}
\begin{tabularx}{\textwidth}{>{\raggedright\arraybackslash}X>{\raggedright\arraybackslash}m{0.105\textwidth}*{2}{C}>{\centering\arraybackslash}m{0.145\textwidth}*{3}{C}}
\toprule
\rowcolor{tableheadergray}
\textbf{Comparison} & \textbf{Dataset} & $\boldsymbol{\bar d}$ & $\boldsymbol{s_d}$ & \textbf{95\% CI} & $\boldsymbol{t(9)}$ & $\boldsymbol{d_z}$ & $\boldsymbol{p_{\mathrm{Holm}}}$ \\
\midrule
DAPO I $-$ Distillation & MetroMap-lite & +20.00 & 4.676 & [16.66, 23.34] & 13.527 & 4.278 & 0.000002207 \\
DAPO I $-$ Distillation & TravelMap-lite & +7.50 & 3.096 & [5.29, 9.71] & 7.661 & 2.423 & 0.000187 \\
DAPO I $-$ Distillation & Fee-VL & +1.80 & 1.199 & [0.94, 2.66] & 4.746 & 1.501 & 0.002102 \\
DAPO I $-$ Distillation & Cancel-VL & +1.19 & 0.880 & [0.56, 1.82] & 4.276 & 1.352 & 0.002102 \\
\lightrowrule
DAPO II $-$ PR immediate & MetroMap-lite & +6.75 & 3.105 & [4.53, 8.97] & 6.875 & 2.174 & 0.000363 \\
DAPO II $-$ PR immediate & TravelMap-lite & +7.25 & 3.438 & [4.79, 9.71] & 6.669 & 2.109 & 0.000367 \\
DAPO II $-$ PR immediate & Fee-VL & +17.50 & 4.799 & [14.07, 20.93] & 11.531 & 3.646 & 0.000007563 \\
DAPO II $-$ PR immediate & Cancel-VL & +5.30 & 3.200 & [3.01, 7.59] & 5.237 & 1.656 & 0.001610 \\
\bottomrule
\end{tabularx}
\caption{Measured paired-test results for DAPO-stage gains. DAPO I is measured relative to Reasoning Distillation; DAPO II is measured relative to the immediate post-reconstruction evaluation. The eight exploratory comparisons form a separate Holm family.}
\label{tab:paired-dapo-gains}
\end{table}

All eight stage gains remain significant after within-family correction. DAPO I has a particularly large effect on MetroMap-lite, whereas DAPO II has its largest VL gain on Fee-VL, consistent with the task-family analysis in the main text. These stage tests are exploratory and are reported separately from the component-level primary family.

\paragraph{Reproducibility records.}
The statistical archive contains 360 measured paired-difference records, 480 measured configuration scores, the unrounded test results, and the scripts used to reproduce the calculations. When one configuration participates in multiple contrasts, the same measured score is reused rather than regenerated. Run seeds and configuration hashes identify the actual experiments. These seed-level paired tests do not replace prediction-level bootstrap or McNemar analyses, which address a different source of uncertainty.

\section{Method Details}
\label{app:detailed-method}

This appendix expands the complete method from end to end. We first present the full training algorithm so that every state transition is visible in one place. We then unpack the state contract and the four operational components---Harness I, Policy Learning I, Harness II, and Policy Learning II---including their inputs, intermediate records, validation conditions, commit rules, and outputs. Task-specific evidence budgets and matched controls are given with the component that uses them.

\subsection{End-to-End Training Algorithm}
\label{app:end-to-end}

Algorithm~\ref{alg:rlharness} is the top-level execution order. Calls to \textsc{ExplorationDistillationHarness} and \textsc{PostRLReconstructionHarness} update only the external program, whereas \textsc{LoRASFT}, \textsc{DAPO}, and \textsc{DAPOResume} update only model parameters. The three exchanged artifacts are the first committed Harness $\Phi^1$, its version-aligned teacher traces $\mathcal G^1$, and the fresh rollout $\mathcal R_{\theta^1}$ produced by the first learned policy. Test data never enters any line of the algorithm.

\begin{algorithm}[ht]
\caption{RLHarness: Complete Training Schedule}
\label{alg:rlharness}
\footnotesize
\begin{algorithmic}[1]
\Require $\mathcal D_{\mathrm{mm}},\mathcal D_{\mathrm{val}},\Psi^0$, base policy $\theta^0$, explorer $S$, teacher $T$, evaluator $Q$, verifier $V$
\Ensure first Harness $\Phi^1$, reconstructed Harness $\Phi^2$, final policy $\theta^2$
\State $(\Phi^1,\mathcal G^1)\gets\Call{ExplorationDistillationHarness}{\mathcal D_{\mathrm{mm}},\mathcal D_{\mathrm{val}},\Psi^0,S,T,V}$
\State $\mathcal D_{\mathrm{SFT}}^1\gets\Call{MaterializeAndAudit}{\mathcal G^1}$ \Comment{data processing outside the Harness}
\State $\theta_{\mathrm{SFT}}\gets\Call{LoRASFT}{\theta^0,\mathcal D_{\mathrm{SFT}}^1,2\text{ epochs}}$
\State $\theta^1\gets\Call{DAPO}{\theta_{\mathrm{SFT}},\Phi^1,0{:}400,R,\text{Hybrid-DGPO}}$
\State $\mathcal R_{\theta^1}\gets\Call{Rollout}{\theta^1,\mathcal D_{\mathrm{mm}},\Phi^1}$ \Comment{fresh current-policy evidence}
\State $\Phi^2\gets\Call{PostRLReconstructionHarness}{\mathcal R_{\theta^1},\mathcal D_{\mathrm{mm}},\Phi^1,\mathcal D_{\mathrm{val}},T,Q,V}$
\State $\theta^2\gets\Call{DAPOResume}{\theta^1,\Phi^2,400{:}600,R,\text{Hybrid-DGPO}}$
\State \Return $\Phi^1,\Phi^2,\theta^2$
\end{algorithmic}
\end{algorithm}

The algorithm contains two asymmetric feedback links. Harness I converts raw task experience into an executable program and aligned supervision before parameter learning. After the policy changes, Harness II reads the new behavior and replaces the learning environment, but does not itself optimize the model. The learned parameter state is $\theta^2$, while the deployed system and final evaluation use the frozen $\theta^2$--$\Phi^2$ pair; $\Phi^1$ and $\Phi^2$ specify what the policy executes at each stage.

\subsection{State Representation and Commit Protocol}
\label{app:harness-boundary}

\paragraph{Task interface and notation.}
For maps, $x=(m,t,q)$ and the target is a complete ordered path $\pi=(u_0,\ldots,u_L)$; the verifier checks endpoints, order, constraints, and the whole route. For VL, $x=(i,r,q)$ contains one frozen task image and the target contains fact-label vector $\mathbf z$, with Cancel-VL additionally predicting responsibility class $c$; the verifier checks fact labels and the responsibility output separately, while the reported primary metric covers only the fact labels. Throughout this appendix, $\mathcal D_{\mathrm{mm}}=\mathcal D_d^{\mathrm{train}}$ is the complete training pool for the current task and $\mathcal D_{\mathrm{val}}$ is frozen before candidate generation. Source inputs, input assembly $I_d$, output contract $C_d$, and verifier $V_d$ remain fixed for all four tasks.

RLHarness stores short-term analysis separately from long-term execution. The short-term context contains only the evidence needed for the current exploration batch, contrastive batch, or repair. The long-term context stores the committed program $\Phi=(I,B,P,E,C)$, of which only $B/P/E$ may receive a new version. New experience first enters the short-term context and is written to persistent state only after it has been compressed into a reusable rule and verified; the raw trajectories are then discarded. Control plane $\Gamma_d$ constructs temporary contexts, invokes the verifier, repairs or rolls back failed revisions, and atomically commits complete versions. It is fixed control logic rather than optimized Prompt content. Context management therefore covers not only token budgeting, but also evidence selection and organization, commit timing, version synchronization, and recovery after failure.

The two Harnesses use the same interface at different points in training. Harness I builds $\Phi^1$ from the original multimodal examples and task Prompt and exports same-version teacher executions $\mathcal G^1$; Harness II starts from fresh post-DAPO-I rollouts and reconstructs $\Phi^2$. SFT-label materialization, SFT, and both DAPO blocks lie outside the Harness boundary: label materialization only changes the data representation, while SFT and DAPO update model parameters. None of the four static tasks requires tool calls or an environment state machine; automation lies in execution dispatch, task verification, repair and retry, and version commitment.

\begin{table}[ht]
\centering
\footnotesize
\renewcommand{\arraystretch}{1.14}
\setlength{\tabcolsep}{3.6pt}
\begin{tabularx}{\textwidth}{>{\raggedright\arraybackslash}m{0.16\textwidth}LL}
\toprule
\rowcolor{tableheadergray}
\multicolumn{1}{c}{\textbf{Aspect}} & \multicolumn{1}{c}{\textbf{Harness I: exploration--distillation}} & \multicolumn{1}{c}{\textbf{Harness II: post-RL reconstruction}} \\
\midrule
Problem & No stable executable program before training & Old program no longer matches the learned policy \\
\lightrowrule
Evidence & A fixed task-specific sample of 480 exploration examples & Fresh rollouts from $\theta^1$ \\
\lightrowrule
Context management & Preassigned batches; task-specific selective commit & Outcome- and difficulty/label-stratified contrasts \\
\lightrowrule
External update & Establish and refine Skills, protocols, and demonstrations & Prune stale content and synchronously rewrite all three \\
\lightrowrule
Output & Harness $\Phi^1$ and aligned traces $\mathcal G^1$ & Reconstructed Harness $\Phi^2$ \\
\lightrowrule
Consumer & SFT and DAPO I & DAPO II \\
\bottomrule
\end{tabularx}
\caption{The two Harnesses solve different problems and consume different evidence. Harness I is task-coverage oriented; Harness II is conditioned on the current policy boundary.}
\label{tab:harness-comparison}
\end{table}

\begin{table}[ht]
\centering
\footnotesize
\renewcommand{\arraystretch}{1.14}
\setlength{\tabcolsep}{4.0pt}
\begin{tabularx}{\textwidth}{>{\raggedright\arraybackslash}m{0.22\textwidth}L>{\raggedright\arraybackslash}m{0.30\textwidth}}
\toprule
\rowcolor{tableheadergray}
\multicolumn{1}{c}{\textbf{Object or stage}} & \multicolumn{1}{c}{\textbf{Input}} & \multicolumn{1}{c}{\textbf{Output and responsibility}} \\
\midrule
Exploration--Distillation Harness &
Original Prompt, complete image--text/table training pool, frozen validation pool, a seeded 480-example exploration subset, explorer, task teacher, and verifier. &
Build the first Harness $\Phi^1$ from the exploration subset and export the audited accepted records $\mathcal G^1$ from complete-training-pool attempts. \\
\lightrowrule
SFT label materialization (outside Harness) &
$\mathcal G^1$ and fixed audit rules. &
Produce frozen labels $\mathcal D_{\mathrm{SFT}}^1$ without changing the Harness or model. \\
\lightrowrule
Policy Optimization I (outside Harness) &
$\Phi^1$, $\mathcal D_{\mathrm{SFT}}^1$, and the base model. &
Run SFT followed by DAPO I to obtain current policy $\theta^1$. \\
\lightrowrule
Post-RL Reconstruction Harness &
Fresh rollouts from $\theta^1$, their original examples, $\Phi^1$, and the frozen validation pool. &
Construct contrastive contexts, reconstruct the external program, and atomically commit $\Phi^2$. \\
\lightrowrule
Policy Optimization II (outside Harness) &
Frozen $\Phi^2$ and checkpoint 400. &
Continue under the same reward and optimizer settings to obtain~$\theta^2$. \\
\lightrowrule
Reward / Hybrid-DGPO / scorer &
Rollouts, references, and task rules. &
Construct policy-learning signals or evaluation results; do not edit the Harness. \\
\bottomrule
\end{tabularx}
\caption{Lifecycle and implementation boundary. Harness I builds the first learnable Harness; Harness II reconstructs it from post-RL behavior. SFT and DAPO update model parameters and are not part of either Harness.}
\label{tab:harness-boundary}
\end{table}

Harness II is run once; after $\Phi^2$ is committed, it remains fixed throughout DAPO II. This separates the effect of Harness reconstruction from the extra budget of repeated search.

\paragraph{State decomposition.}
The full data flow therefore contains three state classes. Trajectories inside an exploration or mixed batch form the \emph{short-term analysis context} and are evicted after that batch. $I/B/P/E/C$ form the \emph{persistent execution context} $\Phi$, but only $B/P/E$ can receive a new committed version. SFT and DAPO weights form the \emph{parameter state} $\theta$. Both Harnesses distill information from short-term context into $B/P/E$ without changing $I/C$ or $\theta$.

\subsection{Exploration--Distillation Harness}
\label{app:harness-one-details}

\paragraph{Overview and interface.}
Before policy learning, Harness I turns a task-only Prompt into a program that a student can execute, a verifier can check, and SFT can imitate. Its outputs are both the inference Prompt $\Phi^1$ and version-aligned teacher traces $\mathcal G^1$:
\[
(\mathcal D_{\mathrm{mm}},\mathcal D_{\mathrm{val}},\Psi^0,S,T,V)
\xrightarrow{\mathcal H_{\mathrm{ED}}}(\Phi^1,\mathcal G^1).
\]
It is neither Skill generation alone nor SFT itself.

\paragraph{Step 1: task-specific program initialization.} Initialization is not shared across task families. For MetroMap-lite and TravelMap-lite, the teacher organizes $\Psi^0$ into exactly five image Skills, five text/table Skills, and five cross-modal fusion Skills, completes the selection rule, execution protocol, and How-to instructions, and inherits and freezes the task output contract. For Fee-VL and Cancel-VL, $\Psi^0$ is already a large, rule-rich Prompt derived from Didi Chuxing's ride-hailing responsibility-assignment logic and adjudication Skills. Harness I preserves its business rules, evidence priorities, and exclusion conditions while organizing them into an executable Skill bank, selection/execution protocol, and How-to instructions, and inherits and freezes the original output contract; it does not force the map-specific $5/5/5$ structure onto the VL tasks. Both paths produce a complete executable program rather than an isolated Skill list.

\paragraph{Step 2: verified demonstration construction.} The fixed demonstration inputs are solved under that initial program: maps use two fixed, distinct training examples and each VL fact label uses one. GPT-5.6-sol supplies map traces and Kimi-K3 supplies VL traces. For the released maps, deterministic checks cover output structure, section order, exact route, selected/used marker consistency, and reproducible route metrics; visual-topology and natural-language execution semantics are not claimed as deterministically verified. The same teacher retries a rejected trace, and only accepted records enter the seed Prompt.

\paragraph{Step 3: fixed-batch exploration.} The task-specific sampler selects 480 examples once and freezes their order. In the released map pipeline, seed 42 samples proportionally by difficulty from Train1600 and partitions the result into twelve ordered batches of 40; the private VL runs use thirty batches of 16. Maps use Gemini-3.5-Flash as explorer and GPT-5.6-sol as teacher, whereas VL uses Kimi-K2.5 and Kimi-K3. The Prompt is frozen within each batch, so all outcomes in that batch are comparable under one external-program version.

\paragraph{Step 4: execution scoring and classification.} Each execution returns the prediction, visible reasoning, and Skill markers. The released map adapter records format validity, prefix partial accuracy, exact-route correctness, and the reference route; the teacher then contrasts successes and failures using these records and their associated task evidence. The private VL pipeline uses its label verifier. These records form the temporary evidence for the current update.

\paragraph{Step 5: candidate synthesis.} The teacher reflects on success and failure minibatches, merges the supported suggestions, and ranks them into one full-Prompt candidate. No recommendation or no effective edit yields a no-op. For the released maps, candidate invariants freeze placeholders, scoring semantics, the Skill-usage contract, output format, and Skill numbering/order; Skill descriptions, ordinary reasoning guidance, and examples may be edited inside the candidate.

\paragraph{Step 6: selective commitment and context eviction.} A structurally valid map candidate is executed by the explorer on the complete frozen Val100 and is committed only when hard route accuracy strictly exceeds the incumbent; ties and decreases are rejected. The two fixed demonstrations are generated and verified when the seed Prompt is created, but the released SkillOpt loop does not separately rerun them after every accepted edit. The batch evidence is then evicted, so only the selected complete Prompt crosses the batch boundary. The private VL implementation uses its corresponding verifier and acceptance test under the same external-state boundary.

\paragraph{Step 7: version-aligned supervision export.} After the last exploration batch, the committed state becomes $\Phi^1$. The teacher attempts every training example under this fixed version. Rejected outputs are retried across at most three generation rounds; the accepted set must pass the exact-response, output-structure and section-order, selected/used-marker, and numerical audits. Only these accepted records form $\mathcal G^1$, so the SFT corpus remains aligned with the Harness later presented to the policy without implying that all 1,600 attempts are retained. The external data pipeline, not Harness I, materializes the accepted records as SFT data.
The resulting flow is
\[
\begin{aligned}
\Psi^0 &\rightarrow\text{initial program and demonstrations}
\rightarrow\text{fixed-batch exploration}\\
&\rightarrow\text{invariant and validation gating}
\rightarrow\Phi^1\rightarrow\mathcal G^1.
\end{aligned}
\]

\subsection{First-Stage Policy Learning}
\label{app:policy-learning-one}

This component converts the first external program into model capability and then produces the evidence needed by Harness II. It changes $\theta$ while keeping $\Phi^1$ fixed throughout both supervised and reinforcement learning.

\paragraph{Step 1: label auditing and materialization.} The external data pipeline reads $\mathcal G^1$, rejects records that fail the fixed format or consistency checks, and serializes the surviving input--trajectory--answer triples as $\mathcal D_{\mathrm{SFT}}^1$. This operation changes data representation only; it does not edit the Harness or~model.

\paragraph{Step 2: execution-trace distillation.} Starting from $\theta^0$, two LoRA-SFT epochs train the policy to reproduce the Skill choice, multimodal evidence binding, intermediate computation, and answer verification demonstrated in $\mathcal G^1$. The executable context remains $\Phi^1$, so the labels and Prompt describe the same program version.

\paragraph{Step 3: first-stage DAPO optimization.} From $\theta_{\mathrm{SFT}}$, DAPO steps $0{:}400$ use the common task reward and Hybrid-DGPO advantage. Only policy parameters and optimizer state change; Skills, protocols, examples, scorers, and source inputs remain fixed.

Both DAPO blocks share $R=0.05r_{\mathrm{format}}+0.25r_{\mathrm{partial}}+0.70r_{\mathrm{all}}$. All four tasks apply rollout-difficulty weighting at the advantage level. Fee-VL and Cancel-VL additionally apply per-fact-label frequency weights $w_l^+=N_l^-/N_l^+$ and $w_l^-=1$ inside the partial-reward computation. The scorer, weights, and advantage definition are unchanged between the two blocks; Appendix~\ref{app:dapo-config} gives the complete specification.

\paragraph{Step 4: policy checkpointing.} Step 400 defines $\theta^1$ and the optimizer state from which DAPO II will later resume. Saving this boundary before reconstruction ensures that the second training phase differs in its Harness rather than in its initialization.

\paragraph{Step 5: current-policy rollout collection.} The complete training pool is rolled out again with $\theta^1$ and $\Phi^1$, producing $\mathcal R_{\theta^1}$. Harness II receives these current predictions, trajectories, Skill-use records, verifier outcomes, and original examples; it does not reuse the stale failures that created $\Phi^1$.

\subsection{Post-RL Harness Reconstruction}
\label{app:harness-two-details}

\paragraph{Overview and interface.}
After DAPO I, the policy has internalized part of the old program and exhibits a new error distribution. Harness II therefore analyzes only fresh rollouts from $\theta^1$, rather than replaying first-stage bad cases, and jointly reconstructs the Skill bank, protocols, \mbox{and demonstrations}:
\[
(\mathcal R_{\theta^1},\mathcal D_{\mathrm{mm}},\Phi^1,
\mathcal D_{\mathrm{val}},T,Q,V)
\xrightarrow{\mathcal H_{\mathrm{PR}}}\Phi^2.
\]
It returns a Prompt and performs no weight update.

\paragraph{Step 1: reconstruction-input freezing.} Harness II receives the complete fresh rollout $\mathcal R_{\theta^1}$ together with the unchanged source examples and $\Phi^1$. Predictions, reasoning records, selected and used Skills, verifier outcomes, and task metadata are aligned at the example level before analysis. Neither the policy nor the Prompt changes while these records are being organized.

\paragraph{Step 2: Skill-usage auditing.} The system counts declared Skill selections and \texttt{[Using Skill N]} markers in $\mathcal R_{\theta^1}$ and retains their union as the observable usage signal. In the released map pipeline, Skills whose selection/marker union appears in less than 0.5\% of Train1600 rollouts become pruning candidates and are removed from the working bank $B^{\prime}$ before candidate synthesis. This is an evidence-based cleanup of the old external program, not a policy update.

\paragraph{Step 3: contrastive evidence construction.} Maps sample a balanced set of successes and failures and preserve difficulty strata; VL constructs a separate pool for every fact label, balances correct and incorrect predictions, and makes gold positives and negatives as even as the available data permits. The selected records are packed into 16-trace mixed batches so each teacher call contains an explicit behavioral contrast.

\paragraph{Step 4: batch-level error diagnosis.} For each batch, the teacher identifies the earliest stage at which failing and successful executions diverge, the multimodal evidence omitted or misused there, procedures already executed reliably, and Skills whose content or ordering no longer matches the policy. The output is a structured diagnosis rather than an edited Prompt.

\paragraph{Step 5: cross-batch diagnosis aggregation.} Diagnoses are accumulated across batches by Skill and reasoning stage. Redundant observations are merged, contradictions remain explicit, and unsupported one-off failures are not promoted. Once a batch has been summarized, its raw traces are evicted; the synthesis call therefore sees compact cross-batch findings rather than an ever-growing rollout transcript.

\paragraph{Step 6: Harness candidate synthesis.} From the same diagnosis set and pruned working bank $B'$, the teacher creates conservative, stage-organized, and compact alternatives. Every candidate is a complete executable package, but may revise only $B/P/E$: the Skill bank (including How-to instructions), selection and execution protocol, and few-shot block. Input assembly $I_d$, output contract $C_d$, source inputs, and verifier remain fixed. Candidate styles change the organization and degree of compression, not the underlying evidence.

\paragraph{Step 7: version-aligned usage and demonstration rewriting.} For each candidate, the teacher rewrites the How-to block, selection lists, Skill markers, and marker-local reasoning from the fixed source demonstrations. The task facts, demonstration identities and order, numerical values, required procedure, and final answers must remain unchanged. The map implementation retains two demonstrations, while Fee-VL and Cancel-VL retain one per fact label. Each rewritten block is validated and inserted only into the candidate that produced it.

\paragraph{Step 8: candidate and few-shot gates.} Candidate generation and demonstration rewriting use separate deterministic gates. The candidate gate checks the Skill schema and verifies that the teacher-produced \texttt{operation\_log} and \texttt{coverage\_audit} process every source Skill consistently. The few-shot gate checks fixed example identities, selection and marker consistency, nonduplicated markers, preserved task facts and numerical values, the required procedure, and final responses. These checks establish structural and record-level consistency; they do not by themselves prove that every natural-language or visual claim is correct. Generation and rewrite calls each retry at most three times, and an unresolved candidate is excluded before Val100 execution.

\begin{table}[ht]
\centering
\scriptsize
\renewcommand{\arraystretch}{1.12}
\setlength{\tabcolsep}{2.8pt}
\begin{tabularx}{\textwidth}{>{\raggedright\arraybackslash}m{0.20\textwidth}LLL}
\toprule
\rowcolor{tableheadergray}
\textbf{Check item} & \textbf{Implementation} & \textbf{Required annotation or evidence} & \textbf{Failure handling} \\
\midrule
Candidate schema and operation coverage & Deterministic parser over Skills, operation log, and coverage audit & Source-Skill operations and dispositions & Return exact error; generation retries up to three times \\
\lightrowrule
Few-shot preservation & Deterministic comparison with the fixed source examples & Identities, facts, values, procedure, and final responses & Reject the rewrite; retry up to three times \\
\lightrowrule
Selection and marker consistency & Deterministic list, set, and marker checks & Selected Skills, used markers, placements, and coverage status & Reject duplicated or mismatched markers \\
\lightrowrule
Candidate execution and selection & Qwen3.5-plus execution and task scoring; task-teacher report selection & Frozen validation labels, metrics, and representative traces & Exclude invalid candidates; commit one complete $B/P/E$ package \\
\bottomrule
\end{tabularx}
\caption{Implemented candidate checks, report generation, and selection roles.}
\label{tab:candidate-checks}
\end{table}

\paragraph{Step 9: candidate execution, reporting, and atomic commitment.} Qwen3.5-plus executes every valid complete candidate on the fixed validation pool and produces task score, format and truncation rates, Skill coverage, use rate, concentration, and representative traces. The task teacher---GPT-5.6-sol in the released map pipeline---then reads the normalized reports and returns the selected candidate and a concise rationale. The checkpoint-400 student is neither candidate executor nor selector: its fresh rollouts supply reconstruction evidence, and it later consumes the committed Prompt during DAPO II. Skill use is examined first, followed by accuracy and stability costs; there is no human ranking, fixed score-gap threshold, or test-set access. The selected complete package is committed atomically as $\Phi^2$.
The resulting flow is
\[
\begin{aligned}
\mathcal R_{\theta^1}&\rightarrow\text{Skill-use audit and pruning}
\rightarrow\text{stratified contrastive batches}
\rightarrow\text{structured diagnoses}\\
&\rightarrow\text{three full candidates}
\rightarrow\text{usage/demonstration rewriting and gating}
\rightarrow\Phi^2.
\end{aligned}
\]

\subsection{Second-Stage Policy Learning}
\label{app:policy-learning-two}

The final component tests whether the policy can learn from the reconstructed program. It uses the same optimization objective as DAPO I and changes only the executable Harness and the continued parameter trajectory.

\paragraph{Step 1: reconstructed-Harness freezing.} Once Harness II commits $\Phi^2$, the complete package---Skills, selection and execution protocols, How-to instructions, demonstrations, and output contract---is frozen. No further candidate search or Prompt edit occurs during DAPO II.

\paragraph{Step 2: checkpoint restoration.} Training resumes from the checkpoint-400 parameters $\theta^1$ and their associated optimizer state. It does not restart from the base or SFT model, so the second phase measures adaptation of the already learned policy to $\Phi^2$.

\paragraph{Step 3: fixed-objective continuation.} DAPO II reuses the same task reward, Hybrid-DGPO advantage, rollout configuration, and scorer as DAPO I. Holding these components fixed makes the reconstructed Harness the intended change in the learning environment.

\paragraph{Step 4: continued policy optimization.} Steps $400{:}600$ execute examples with $\Phi^2$ and update only model parameters, yielding $\theta^2$. The Harness remains readable external state; the policy is responsible for internalizing and executing its revised trigger boundaries, order, and demonstrations.

\paragraph{Step 5: held-out evaluation.} Held-out test examples are used for final measurement of the frozen $\theta^2$--$\Phi^2$ pair. They are excluded from reconstruction, candidate validation, and every \mbox{optimization decision}.

\subsection{Evidence Budgets and Matched Controls}
\label{app:harness-evidence-controls}

For the map tasks, fixed seed 20260917 selects 128 checkpoint-400 traces: 64 successes and 64 failures, stratified within each group by Map Difficulty $\times$ Query Difficulty. They are packed into eight 16-trace batches, each containing eight successes and eight failures, so GPT-5.6-sol can directly contrast behavior to preserve with failure patterns to repair. Only structured diagnoses survive each batch.

\paragraph{Map-task evidence-budget selection.}
We compared three strictly nested evidence budgets---64, 128, and 256 cases---on the same TravelMap-lite
checkpoint-400 rollout pool. Every budget keeps a 1:1 success--failure ratio
and produces one conservative and one stage-based candidate. All six prompts are evaluated with identical decoding and scoring on a separately frozen 400-example development pool, Budget-Val400, sampled from the training source. Budget-Val400 is sample-ID-disjoint from the rollout evidence used for candidate construction and from Test400; Test400 is not accessed during budget selection.

\begin{table}[ht]
\centering
\footnotesize
\renewcommand{\arraystretch}{1.08}
\setlength{\tabcolsep}{7.0pt}
\begin{tabularx}{\textwidth}{*{6}{C}}
\toprule
\rowcolor{tableheadergray}
\textbf{Cases} & \textbf{Conservative} & \textbf{Stage-based} & \textbf{Mean} & \textbf{Gap} & \textbf{Batches} \\
\midrule
64  & 49.00 & 47.25 & 48.13 & 1.75 & 4 \\
128 & 49.75 & \textbf{49.50} & \textbf{49.63} & \textbf{0.25} & 8 \\
256 & \textbf{50.50} & 47.25 & 48.88 & 3.25 & 16 \\
\bottomrule
\end{tabularx}
\caption{TravelMap-lite evidence-budget audit on training-source Budget-Val400 (Accuracy/\%). Conservative and Stage-based denote the two candidate-construction styles; Batches counts 16-trajectory teacher-analysis batches.}
\label{tab:evidence-budget}
\end{table}

Although the conservative candidate at 256 cases reaches the best single score of 50.50\%, higher than the 128-case result, 128 cases give the highest
mean across candidate styles, the smallest candidate gap, and half as many
teacher-analysis batches as 256. We therefore use 128 as an empirical
map-task default, not as a theoretically optimal constant. Because small differences on Budget-Val400 correspond to only a few examples, this audit is used to illustrate the tradeoff among coverage, noise, and cost rather than to support a significance claim.

Fee-VL and Cancel-VL follow the same budget-selection principle: we seek a
tradeoff among fact-label coverage, stability across candidate styles, and
teacher-analysis cost rather than the best score of one candidate. Because
both tasks require an independent evidence pool for every fact label,
repeating the full budget-sensitivity analysis label by label adds no new
methodological information. We therefore omit those intermediate comparisons
and report only the frozen per-label protocol.

\paragraph{Per-label evidence protocol for vision--language tasks.}
Fee-VL and Cancel-VL construct evidence pools independently for each fact label. Each label receives 64 trajectories: 32 on which that label is predicted correctly and 32 on which it is predicted incorrectly. Within both groups, gold positives and negatives are balanced as closely as the available data permits. The 64 cases are packed directly into four mixed batches of 16, each containing eight correct and eight incorrect predictions and preserving gold-label balance when possible. The 32/32 correctness quota remains fixed; when a class is scarce, only the within-group positive/negative ratio is adjusted to the closest attainable balance and the realized composition is recorded. Label pools are sampled independently, so one multilabel rollout may appear in several pools; $9\times64$ for Fee-VL and $7\times64$ for Cancel-VL count per-label evidence slots rather than unique trajectories.

Map Val100 is frozen before optimization and is used both for the initial strict commit gate and for post-RL candidate selection. Fee-VL and Cancel-VL freeze 50 validation cases per fact label, balancing gold positives and negatives as closely as possible.

For each candidate, the task teacher rewrites the How-to block and fixed demonstrations while preserving their identities, task facts, numerical values, required procedure, and final answers. The map tasks retain two demonstrations, whereas Fee-VL and Cancel-VL retain one per fact label. Candidate-generation and few-shot outputs pass the deterministic checks in Table~\ref{tab:candidate-checks}; the calls retry at most three times, and an unresolved candidate is barred from validation. Qwen3.5-plus executes the valid candidates and produces normalized reports. The task teacher then selects from those reports and Prompts, examining Skill use first and accuracy, format, truncation, and stability costs afterward. The checkpoint-400 student does not run candidate validation, and neither test data nor human ranking enters the decision.

The minus-Few-shot control keeps the selected Post-RL Skill Bank, its How-to instructions, the checkpoint-400 policy, and every evaluation setting fixed, but removes the Few-shot demonstration block from the executable prompt. It does not replace the Post-RL Reconstruction Harness or its diagnoses. The comparison therefore isolates whether examples are needed for the current policy to recognize and execute the revised harness.

Two matched controls isolate context management. \emph{Minus Selective Context Commit} keeps the same fixed 480 first-stage examples, their order, task-specific batch plan, models, verifier, call budget, and edit operators fixed but disables selective rejection of a legal effective candidate. \emph{Minus Stratified Contrastive Evidence Curation} keeps the second-stage evidence volume, number of 16-shot batches, teacher, three-candidate synthesis, few-shot reconstruction, validity gate, validation pool, and selector fixed. Maps instead sample 128 rollouts uniformly and randomly pack $8\times16$ without correctness or difficulty quotas; VL samples 64 per label and randomly packs $4\times16$ without correctness or gold-label balancing. These controls change persistence and evidence composition, respectively, not the evidence or compute budget. Table~\ref{tab:joint-ablation} reports the stratified-evidence branch both immediately after reconstruction and after its matched DAPO-II continuation.

\subsection{Stage-Specific Algorithms}
\label{app:stage-algorithms}

Algorithms~\ref{alg:harness-one} and~\ref{alg:harness-two} turn the detailed prose into executable stage boundaries. The top-level ordering is already given by Algorithm~\ref{alg:rlharness}. \textsc{Verify} supplies task feedback for one execution, while \textsc{CandidateGate} and \textsc{FewShotGate} validate the two generated artifacts before candidate execution; none of these operations ranks candidates.

\begin{algorithm}[H]
\caption{Harness I: Exploration--Distillation}
\label{alg:harness-one}
\footnotesize
\begin{algorithmic}[1]
\Require training pool $\mathcal D_{\mathrm{mm}}$, validation pool $\mathcal D_{\mathrm{val}}$, original Prompt $\Psi^0$
\Statex \hspace{\algorithmicindent}\textbf{Models:} explorer $S$, task teacher $T$, verifier $V$
\Ensure first Prompt $\Phi^1$, version-aligned teacher traces $\mathcal G^1$
\State $\Phi\gets\Call{TaskSpecificInitialize}{T,\Psi^0}$ \Comment{map: $5/5/5$; VL: rule-rich Prompt}
\State $E^0\gets\Call{SolveAndVerifyDemonstrations}{T,\Phi,V}$
\State $\Phi\gets\Call{Assemble}{\Phi,E^0}$
\State $(\mathcal D_{\mathrm{ED}},\mathcal Q)\gets\Call{TaskSpecificBatchPlan}{\mathcal D_{\mathrm{mm}},480}$
\Statex \hspace{\algorithmicindent}\textit{Map release: seed 42 and $12\times40$; VL: $30\times16$}
\State $\rho\gets\Call{InitialCommitScore}{S,\Phi,\mathcal D_{\mathrm{val}},V}$
\ForAll{$\mathcal B_t\in\mathcal Q$}
    \State $Z_t\gets\{\Call{Execute}{S,x,\Phi}:x\in\mathcal B_t\}$ \Comment{one Prompt version per batch}
    \State $F_t\gets\{\Call{Verify}{V,x,z}:(x,z)\in(\mathcal B_t,Z_t)\}$
    \State $\Delta_t\gets\Call{ReflectMergeRank}{T,\Phi,\mathcal B_t,Z_t,F_t}$
    \State $\widetilde\Phi\gets\Call{ApplyCandidate}{\Phi,\Delta_t}$
    \State $\widetilde\rho\gets\Call{CandidateScore}{S,\widetilde\Phi,\mathcal D_{\mathrm{val}},V}$
    \If{$\Call{AcceptTaskSpecific}{\widetilde\Phi,\widetilde\rho,\rho,V}$}
        \State $\Phi\gets\widetilde\Phi$; $\rho\gets\widetilde\rho$ \Comment{map: strict Val100 gain}
    \EndIf
    \State $\Call{Evict}{\mathcal B_t,Z_t,F_t}$
\EndFor
\State $\Phi^1\gets\Phi$; $\mathcal G^1\gets\varnothing$
\ForAll{$x\in\mathcal D_{\mathrm{mm}}$}
    \State $g\gets\Call{GenerateAndAudit}{T,x,\Phi^1,V}$ \Comment{at most three rounds}
    \If{$g$ is accepted}
        \State $\mathcal G^1\gets\mathcal G^1\cup\{g\}$
    \EndIf
\EndFor
\State \Return $\Phi^1,\mathcal G^1$
\end{algorithmic}
\end{algorithm}

\begin{algorithm}[H]
\caption{Harness II: Post-RL Reconstruction}
\label{alg:harness-two}
\footnotesize
\begin{algorithmic}[1]
\Require rollouts $\mathcal R_{\theta^1}$ and Prompt $\Phi^1$
\Statex \hspace{\algorithmicindent}\textbf{Pools:} examples $\mathcal D_{\mathrm{mm}}$, validation $\mathcal D_{\mathrm{val}}$
\Statex \hspace{\algorithmicindent}\textbf{Models:} teacher/selector $T$, candidate executor $Q$, verifier $V$
\Ensure reconstructed Prompt $\Phi^2$
\State $B\gets\Call{Bank}{\Phi^1}$; $B'\gets\Call{PruneByObservedUsage}{\mathcal R_{\theta^1},B}$
\State $\mathcal Q\gets\Call{TaskSpecificStratifiedPack}{\mathcal R_{\theta^1},\mathcal D_{\mathrm{mm}},16}$
\State $A\gets\varnothing$
\ForAll{$q\in\mathcal Q$}
    \State $a_q\gets\Call{CompareAndDiagnose}{T,q,B'};\ A\gets A\cup\{a_q\};\ \Call{Evict}{q}$
\EndFor
\State $\mathcal C\gets\Call{GenerateFullPromptCandidates}{T,A,\Phi^1,B'}$ \Comment{three fixed styles}
\ForAll{$c\in\mathcal C$}
    \State $c\gets\Call{RewriteUsageAndDemonstrations}{T,c}$
    \State $c\gets\Call{RepairAtMostThreeTimes}{c,\{\mathrm{CandidateGate},\mathrm{FewShotGate}\}}$
    \If{$c$ is valid}
        \State $\rho_c\gets\Call{ExecuteAndReport}{Q,c,\mathcal D_{\mathrm{val}}}$ \Comment{$\theta^1$ is not run here}
    \Else
        \State remove $c$ from $\mathcal C$
    \EndIf
\EndFor
\State $\Phi^2\gets\Call{SelectFromReports}{T,\mathcal C,\{\rho_c\}}$ \Comment{no test data or human ranking}
\State \Return $\Phi^2$
\end{algorithmic}
\end{algorithm}

Task-specific teachers, explorers, evidence budgets, demonstration counts, and validation-pool construction follow the preceding subsection. Label materialization, SFT, and both DAPO calls remain outside the two Harnesses. The components exchange $\mathcal G^1$, $\mathcal R_{\theta^1}$, and $\Phi^2$, but their implementation boundaries remain explicit.

\clearpage
\section{Prompt Specification and Case Studies}
\label{app:prompt-cases}

This appendix first gives an abbreviated view of the final MetroMap-lite Prompt and then presents one four-stage map case and one Fee-VL Prompt-revision case. The former shows how a topology error moves through Harness I, DAPO I, Harness II, and DAPO II. The latter compares the rule before and after reconstruction on an anonymized original trajectory image. Fee-VL order identifiers, raw record fields, and unnecessary dialogue are omitted.

\subsection{MetroMap-lite Prompt Specification}
\label{app:metromap-prompt}

The final Prompt is not a continuous stack of Skill text, but a versioned program comprising a task contract, scoring rules, Skill Bank, invocation protocol, stage order, demonstrations, and output constraints. The excerpt retains one representative item from each block and uses ellipses for the remainder; the complete version used in the experiments remains fixed throughout evaluation.

\begin{promptbox}[Abbreviated MetroMap-lite Prompt]
# Role
You are a subway path-planning algorithm. Use the attached metro map
and Vertex Table to find the legal route with minimum weighted cost.

# Input Task
{question}

# Weights
- Time: {w1}; Price: {w2}; Comfort: {w3}; Reliability: {w4}

# Input Data
One metro-map image and one Vertex Table are provided.

# Authoritative Scoring Rules
Score the start and all intermediate stations, but exclude the destination.
...

# Planning Skills
1. Build a Map-Grounded Route Sequence
   Trigger: when constructing any route segment.
   Action: follow visible tracks station by station; never infer adjacency
   from table order or real-world memory.
   Guard/Check: every consecutive pair must be a visible edge.
...

# How to Use the Skills
Select only triggered Skills; mark each Skill once when its Action executes.
...

# Required Procedure
[Task and Map Analysis] -> [Relevant Strategy Selection] ->
[Candidate Routes] -> [Score Calculation] -> [Decision and Verification]
...

# Few-Shot Demonstrations
Example 1: one legal route; Example 2: compare multiple candidates.
...

# Output Format Constraints
Output exactly <reasoning>...</reasoning><response>route</response>.
...
\end{promptbox}

Three version boundaries must survive compression. First, task facts and authoritative scoring rules are immutable. Second, a Skill's Trigger, Action, Guard, and Check jointly define its invocation boundary. Third, demonstrations must be rewritten and checked under the same Skill and protocol version; an old example cannot simply be appended to a revised Prompt.

\subsection{Case Studies}
\label{app:end-to-end-cases}

\subsubsection{MetroMap-lite Route-Planning Case Study}
\label{app:metromap-case}

\begin{figure}[H]
\centering
\includegraphics[width=0.92\textwidth]{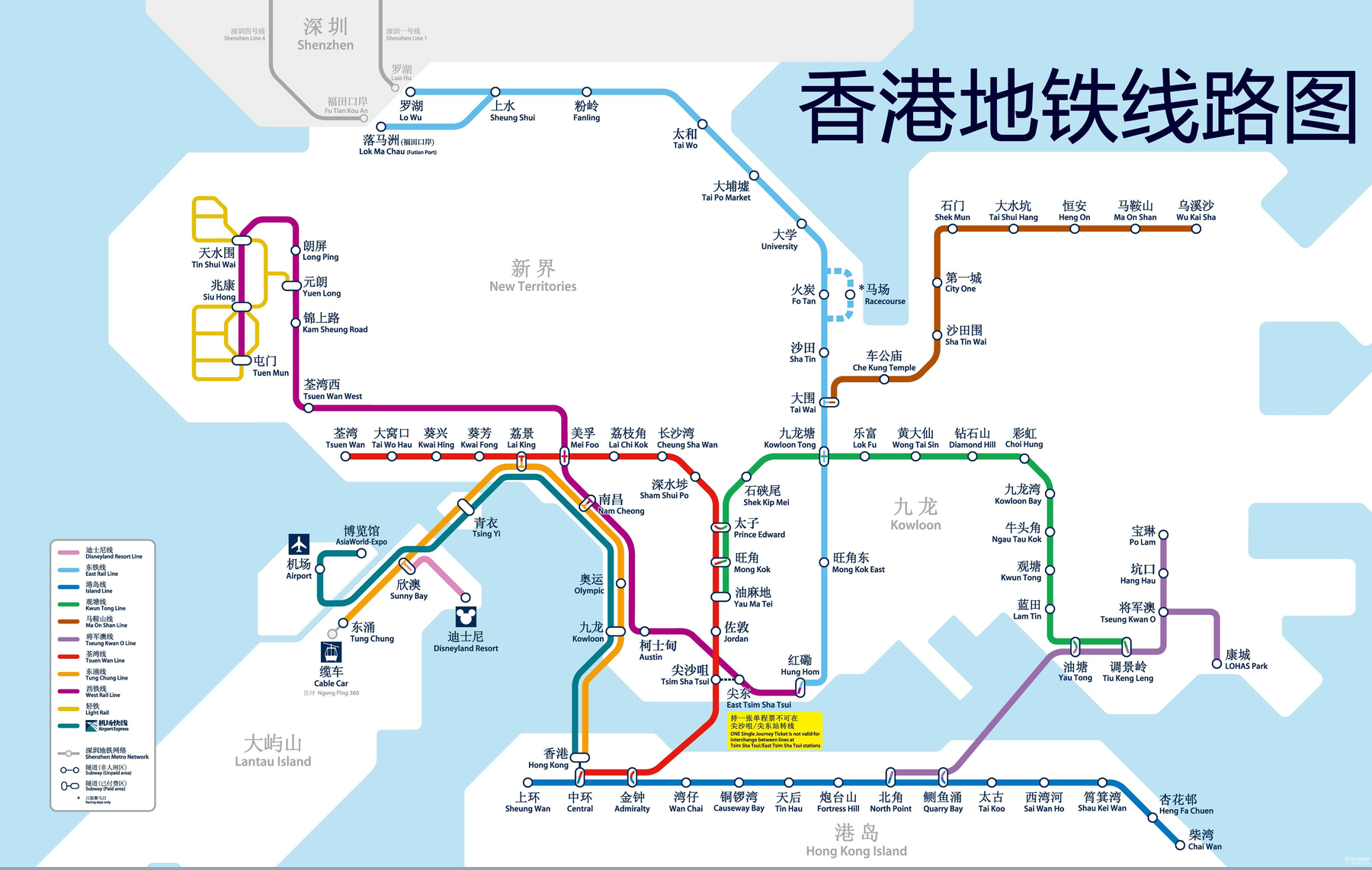}
\caption{Multiline MetroMap-lite input used in the case. The map contains more than five lines; the analysis follows only visible map edges and does not use real-world transit knowledge.}
\label{fig:metromap-workflow-case}
\end{figure}

This case is drawn from the checkpoint-400 reconstruction evidence. Given weights 0.409331/0.153055/0.130721/0.306893 for time, price, comfort, and reliability, the task asks for the minimum-cost legal route from Chai Wan to Tai Koo. The policy output is Chai Wan--Shau Kei Wan--Sai Wan Ho--Tai Koo, whereas the verified route is Chai Wan--Heng Fa Chuen--Shau Kei Wan--Sai Wan Ho--Tai Koo. We place this real error in the complete Harness I--DAPO I--Harness II--DAPO II lifecycle. The final step is a verifier-accepted replay of the behavior required by the revised Prompt, not an additional quantitative result in the main table.

\paragraph{Harness I: initial execution environment.}
The initial Harness makes the image the sole source of adjacency, limits the Vertex Table to canonical names and attributes, and fixes the scoring scope to the start and intermediate stations while excluding the destination. It further requires station-by-station candidate construction, transfer cost only at an actual line change, and comparison under one formula. Teacher-verified demonstrations materialize these rules in the five-stage format used for~SFT.

\paragraph{DAPO I: topology-error propagation.}
The first RL policy produces valid syntax, identifies the requested endpoints, and expands in the correct direction on the Island Line. In this rollout, however, it connects Chai Wan directly to Shau Kei Wan and skips visible Heng Fa Chuen. Once this illegal edge enters the candidate, attribute lookup, weighted scoring, and format checking cannot restore the missing station; exact-route accuracy is therefore zero. The earliest error is topology discretization, not arithmetic.

\begin{promptbox}[DAPO-I output]
Output: Chai Wan-Shau Kei Wan-Sai Wan Ho-Tai Koo
Verifier: incorrect; the first edge skips visible Heng Fa Chuen.
\end{promptbox}

\paragraph{Harness II: topology-guard reconstruction.}
Reconstruction does not add another scoring formula. It routes repeated skipped-node errors to topology construction and inserts a No-Skip Edge Guard before any candidate is scored: every adjacent output pair must be connected by one direct visible edge; if another marker lies on that segment, the model must reread the local crop and expand the sequence. Few-shot trajectories are rewritten under this rule so that the new Prompt no longer demonstrates a brittle one-pass reading strategy.

\begin{promptbox}[Local Prompt revision]
- Trace a visible line from start to destination and check adjacency.
+ After proposing each candidate edge, reread that local track segment.
+ If another visible station lies between the two recorded stations, the
   edge is illegal; insert the station before continuing.
+ Lock and verify the complete station sequence before table lookup or scoring.
\end{promptbox}

\paragraph{DAPO II: adaptation to the reconstructed Harness.}
The second RL block fixes the revised Prompt and retains the same route and format verifier. In replay, the policy first locks the local sequence Chai Wan--Heng Fa Chuen--Shau Kei Wan--Sai Wan Ho--Tai Koo, confirms that no line change occurs, and only then scores the start and intermediate stations. The improvement is not case memorization: the reusable action is to test every proposed edge for an unrecorded station.

\begin{promptbox}[DAPO-II replay]
Output: Chai Wan-Heng Fa Chuen-Shau Kei Wan-Sai Wan Ho-Tai Koo
Verifier: correct; endpoints, edgewise adjacency, order, transfer state,
and output syntax all pass.
\end{promptbox}

The four stages expose the division of state: the Harness stores an auditable topology-reading and verification program, while RL makes the policy invoke it reliably. A Prompt edit alone does not ensure that an old checkpoint executes the new guard, and RL alone does not identify where the guard should be inserted.

\subsubsection{Fee-VL Causal-Attribution Case Study}
\label{app:fee-navigation-case}

\begin{figure}[H]
\centering
\includegraphics[width=0.66\textwidth]{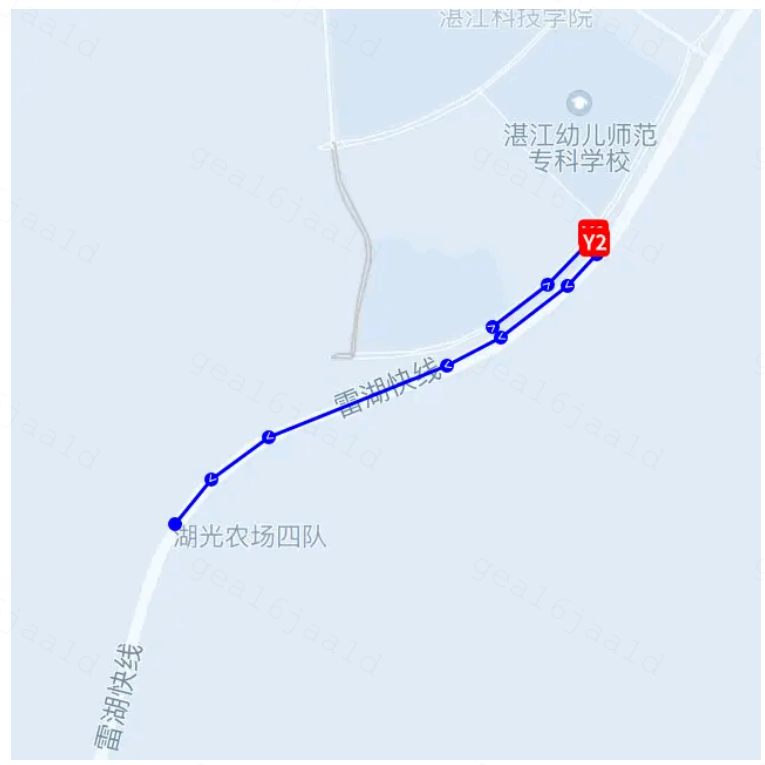}
\caption{Original anonymized trajectory image for the Fee-VL case. It verifies whether a spoken direction became an observed action; it does not directly reveal the responsibility label.}
\label{fig:fee-navigation-timeline}
\end{figure}

\paragraph{Task definition and evidence.}
The anonymized question asks whether the substantial detour was caused by erroneous platform navigation or a map bug. Estimated/actual durations are 5/23 minutes, estimated/actual distances are 14.24/22.90 km, and the endpoint is about 1.41 km from the destination. A ``U-turn'' instruction at 09:45:56 precedes yaw 1 at 09:45:59; yaw 2 occurs around 09:46:19 near replanning; and another ``U-turn'' at 09:50:40 precedes yaw 3 at 09:50:58. Passenger route discussion appears near yaws 1 and 2. The gold label is zero: the evidence does not establish platform-navigation responsibility.

\paragraph{Pre-reconstruction Prompt.}
The original logic aligns yaw anchors, navigation speech, and nearby dialogue, then lets trajectory--announcement agreement support a navigation anomaly.

\begin{promptbox}[Original navigation rule]
Goal: determine whether a detour was caused by an erroneous navigation
announcement or map bug by comparing the driven trajectory with speech.

1. Extract each yaw anchor from the static trajectory.
2. Read navigation audio in the minute before each yaw and driver--passenger
   dialogue within one minute before and after it.
3. Case A (no announcement): infer silent driving and a navigation-planning
   error; treat the trajectory as consistent with navigation.
4. Case B (announcement): if announced direction agrees with the trajectory,
   treat the trajectory as consistent with navigation.
5. Case C (attribution): the driver explicitly blames navigation or the map
   and the passenger does not refute the claim.

Hit: Case A or B, together with Case C.
No hit: trajectory and announcement disagree and there is no explicit claim.
\end{promptbox}

\paragraph{Failure analysis.}
The model correctly observes ``U-turn announcement before yaw -> image-confirmed U-turn -> the driver followed navigation,'' but then turns execution agreement into ``navigation caused the detour'' and produces a false positive. The earliest unrecoverable decision is not a visual error; it treats executing an instruction as sufficient evidence that the instruction was erroneous and causal, while failing to admit passenger direction as a more direct competing explanation.

\paragraph{Post-reconstruction Prompt.}
The Harness completes only this causal boundary; it neither changes the label contract nor stores the case answer.

\begin{promptbox}[Revised navigation rule]
Goal: determine whether the detour was actually caused by erroneous
navigation or a map bug rather than a human route change.

1. Adjudicate every yaw independently.
2. Read navigation audio before the yaw and nearby dialogue; also search for
   passenger directions, destination changes, and voluntary rerouting.
3. No announcement means missing navigation evidence, not a navigation error.
4. Agreement proves only execution. Check whether the instruction conflicts
   with a reasonable route and whether execution caused the yaw or detour.
5. An explicit, unrefuted driver claim is strong navigation-defect evidence.
6. Do not attribute to navigation when a human change explains the yaw more
   directly. Post-yaw replanning cannot cause the earlier yaw.
7. Use the static trajectory to verify execution of directional ASR.

Hit only when execution agrees with navigation, the navigation defect is
explicitly supported, the causal chain holds, and no more direct cause exists.
\end{promptbox}

\paragraph{Corrected decision analysis.}
The rerun still detects coherent navigation speech around all three yaws, and the trajectory confirms that the driver executed directional instructions, but the revised Prompt limits this fact to intermediate evidence. No driver explicitly identifies a navigation or map defect; passenger route discussion supplies a competing explanation near yaws 1 and 2; and post-yaw replanning cannot retroactively cause the earlier yaw. The conjunction of instruction error, causal effect, and absence of a more direct cause therefore fails. The rerun returns no hit, matching the gold label. The important change is from correlation to a temporally directed causal and exclusion test, not simply more Prompt text.

\section{Discussion, Limitations, and Future Directions}
\label{app:discussion}

\subsection{Discussion}
\label{app:interpretation}

The final scores show that staged Harness--RL coordination is effective but far from saturated: MetroMap-lite, TravelMap-lite, Fee-VL, and Cancel-VL reach 62.00\%, 50.00\%, 65.81\%, and 65.51\%. Relative to the strongest external references, the 9B policy still trails by 16.00\%, 14.75\%, and 3.67\% on MetroMap-lite, TravelMap-lite, and Cancel-VL. Both trained checkpoints exceed their matched unadapted baselines on transfer tasks, but the gains are moderate and remain below within-task final performance. The remaining gap is therefore not one missing Prompt rule; it reflects perception, long-horizon credit assignment, task semantics, and optimization budget together.

\paragraph{Visual perception bottlenecks.}
Map tasks require a high-resolution diagram to be discretized reliably into stations and edges, while business VL requires static trajectories, local markers, ASR, and OCR to be aligned on one timeline. A Prompt can specify where to look and in what order, and RL can internalize some shared perception, but neither can recover evidence that was cropped, overlooked, or misrecognized. One omitted station or misread action anchor makes later formulas and responsibility rules ineffective. This explains why structurally similar maps transfer somewhat without producing a dramatic jump.

\paragraph{Sparse long-horizon rewards.}
MetroMap-lite and TravelMap-lite use strict complete-route accuracy: one station, order, endpoint, or transfer error makes the entire answer incorrect. Fee-VL and Cancel-VL use multilabel macro F1, where rare positives, adjacent responsibility classes, and class imbalance concentrate high-variance updates in few examples. Hybrid-DGPO improves difficulty and class weighting, but the terminal reward still does not precisely identify whether the failure arose in perception, candidate construction, arithmetic, or label attribution. DAPO II must therefore learn a long execution chain from relatively coarse feedback.

\paragraph{Harness--policy version coupling.}
Harness II changes Skill triggers, procedure order, and demonstrations after checkpoint 400 has adapted to $\Phi^1$. Immediate replacement by $\Phi^2$ lowers MetroMap-lite/Fee-VL from 57.50\%/49.23\% to 55.25\%/48.31\%; only after the matched 200-step DAPO-II budget do they reach 62.00\%/65.81\%. The revised Harness is a new learning environment, not a cost-free hot patch. Cross-task Prompt replacement is likewise clearly weaker than the task-matched Prompt, showing that triggers, constraint order, and label semantics remain jointly conditioned on task and policy.

\paragraph{Stage-specific roles of the two Harnesses.}
Harness I establishes a common execution language among images, tables, rules, and outputs for an unadapted model and produces complete trajectories for SFT. Harness II operates after one RL block and uses fresh rollouts to separate stable steps, systematic failures, and mismatch between the old Prompt and current behavior. They share selection, organization, compression, verification, recovery, and commitment, but the first emphasizes coverage and cold start whereas the second emphasizes contrastive diagnosis and boundary reconstruction. Collapsing them into one generic Skill iteration would weaken both functions.

\paragraph{Capacity, data, and iteration constraints.}
The student is a 9B model and the study contains only two policy-learning blocks and two Harness versions. Although the reported comparison families include $K=10$ matched runs, the study does not yet provide long-term per-example trajectories across additional update cycles. Evidence budgets of 64/128/256 can only approximate the current error distribution, and teacher-generated candidates need not contain the globally optimal external program. The current results therefore establish the value of staged coordination, not that the method has reached its ceiling under the available data and compute.

\subsection{Limitations}
\label{app:limitations}

\paragraph{Statistical evidence and reproducibility.}
Appendix~\ref{app:statistical-tests} reports $K=10$ measured matched runs: all 24 primary component comparisons are significant after Holm correction, while the four Prompt-replacement and eight DAPO-stage comparisons are exploratory families corrected separately. These tests cover the stated aggregate comparisons, not every label-wise difference or every path through the candidate generator. The reported MetroMap-lite score of 62.00 comes from an earlier reconstructed-Harness branch. It supports external revision followed by re-adaptation but does not by itself establish end-to-end reproduction of the complete current candidate-generation procedure. The four-stage MetroMap-lite narrative in Appendix~\ref{app:metromap-case} joins a real checkpoint-400 error to a verifier-accepted target replay as a mechanism example; the single case is not counted as an additional quantitative result.

\paragraph{Direct method reference.}
SkillRL is the closest Skill-learning reference for our setting. We run it on all four tasks using the original implementation and complete configuration, changing only the datasets and necessary task-interface adapters. Table~\ref{tab:model-comparison} shows that RLHarness outperforms SkillRL on all four tasks, supporting the benefit in our setting of reconstructing a version-aligned Skill Bank, invocation protocol, and demonstrations and then adapting the policy that consumes them.

\paragraph{Why additional Harness baselines are not included.}
We do not add further training-free Harness systems for three reasons. First, our external comparison asks whether system-level adaptation can bring a small open-weight model close to strong closed models. For this capability question, representative closed-model scores provide a common reference in place of a collection of heterogeneous Harness results, as in HarnessForge-style evaluations; they are not evidence that RLHarness is better than the closed systems~\citep{chen2026harnessforge}. Second, existing Harness methods report no results on MetroMap-lite, TravelMap-lite, Fee-VL, or Cancel-VL, so there are no published numbers that can be compared directly. Third, the research objects differ: many Harness methods are training-free and optimize prompts, context organization, or control logic at inference while keeping the task model fixed. They are not designed to test whether a program reconstructed from a changed policy can be synchronized across $B/P/E$ and then internalized by a small model through subsequent SFT or RL. Porting them would require new training objectives, multimodal interfaces, verifiers, and compute budgets, producing a new hybrid rather than a faithful matched baseline. Our evidence therefore establishes gains over the implemented SkillRL baseline and addresses post-update program reconstruction and policy adaptation; it does not rank RLHarness against unimplemented Harness optimization systems.

\paragraph{Remaining comparison and benchmark boundaries.}
Candidate validation uses Qwen3.5-plus as a task-matched proxy; because checkpoint 400 does not execute the candidates before selection, the study does not isolate how candidate rankings would change under direct student execution. Transfer evaluation covers two map tasks and two business-VL tasks and does not establish transfer across more distant task families. Map claims are further limited to MetroMap-lite and TravelMap-lite and therefore exclude the extreme-length examples removed by the lite construction.

\paragraph{Privacy and observability.}
Fee-VL and Cancel-VL expose only aggregates, necessary numerical evidence, and de-identified trajectory images. Teacher analysis of visible trajectories is operational attribution, not causal identification of hidden neural states. When ASR, OCR, or the static image is missing or misread, a rule update cannot recover evidence that was absent or perceived incorrectly.

\paragraph{Scalability and computational cost.}
Both Harnesses rely on strong teachers, verifiers, and candidate evaluation. Larger Skill Banks increase context, synchronization, and validation costs and may amplify teacher bias. The study contains only two policy-learning blocks and two Harness versions; it does not test merging, forgetting, version conflicts, or safety rollback over a long-running~lifecycle.
\subsection{Future Directions}
\label{app:future-work}

\paragraph{Scaling to full-featured Harness--policy systems.}
The present study maintains Skills, selection and execution protocols, and few-shot demonstrations, and therefore evaluates a relatively compact Harness--RL loop. A more mature Harness could additionally incorporate retrieval, tool routing, long-term memory, planning, process verification, failure recovery, and permission or safety constraints as composable, replaceable, and reversible external modules. When combined with RL, the research question is not merely whether adding more Prompt content or tools improves performance, but which capabilities should remain in auditable external modules, which should be internalized in model parameters, and how a policy should learn when to invoke, skip, verify, or recover each module. Harness updates would likewise require credit signals from the current policy's on-policy behavior to distinguish program-design defects from policy-execution failures. Future evaluations should jointly measure task performance, sample and call efficiency, inference latency, version stability, cross-task reuse, and safe rollback, establishing whether a larger Harness genuinely improves the learning environment rather than merely increasing context and system complexity.

\paragraph{Explicit intermediate representations for visual evidence.}
For map tasks, a dedicated parser can first produce a confidence-scored station--edge graph, leaving candidate search and weighted comparison to the language model; low-confidence edges trigger local cropping, enlargement, and a second visual pass. Business VL can first bind trajectory, ASR, and OCR evidence to a shared timeline and order referent while recording both observations and uncertainty. This separates upstream perception failure from downstream rule failure instead of asking one Prompt to perform both recognition and adjudication.

\paragraph{Stage-level rewards and executable verification.}
Map tasks can separately reward endpoints, edgewise adjacency, candidate completeness, transfer state, scoring scope, and the final route. Business VL can verify time windows, evidence referents, observed action, causal direction, and exclusion conditions. Stage rewards do not replace the terminal metric, but shorten its credit path. Formula and time calculations should be delegated to deterministic code so that correct perception is not lost to arithmetic errors.

\paragraph{Failure routing before Harness revision.}
The update interface should separate five destinations: visual uncertainty invokes crop rereading, missing facts invoke retrieval, numerical errors invoke a calculator, isolated execution lapses remain policy-learning targets, and only repeated observable boundary failures authorize Harness editing. Candidate commitment should rerun successful cases and difficulty slices so that a local rule does not damage already stable behavior.

\paragraph{Shared capabilities and task-specific decision boundaries.}
Both trained checkpoints acquire some transferable visual perception, and structurally similar map tasks share more readily, but the gains are not large. VL labels, evidence priority, and responsibility semantics transfer less directly. Future work can share image reading, temporal alignment, candidate retention, and evidence composition through Skills or an intermediate representation while retaining task-private triggers, label semantics, demonstrations, and output contracts; multitask training or parameter-efficient routing can decide which components enter shared weights.

\paragraph{Extended auditable lifecycles.}
Matched-seed repetitions should preserve per-checkpoint predictions, Prompt hashes, candidate provenance, and Gate decisions. Longer Harness--policy cycles should measure forgetting, version compatibility, rollback cost, and safety constraints. Only such records can distinguish a capacity ceiling from inadequate evidence budgets, teacher candidate coverage, or optimization instability.

\section{Acknowledgments}
\label{app:acknowledgments}

This work was sponsored by the CCF-DiDi GAIA Collaborative Research Funds for Young Scholars. The authors gratefully acknowledge the program's support for this collaborative research.

\end{document}